\documentclass[11pt]{article}

\usepackage{arxiv}

\usepackage{times}
\usepackage{latexsym}
\usepackage{natbib}

\usepackage[utf8]{inputenc}
\usepackage{url}
\usepackage[hidelinks]{hyperref} 

\usepackage{graphicx}
\graphicspath{{Figures}}
    \DeclareGraphicsExtensions{.png,.pdf}
\usepackage{subcaption}
\usepackage{multirow}
\usepackage{booktabs}

\usepackage{float}
\usepackage{longtable}

\usepackage{amsmath}

\title{\large Representational Similarity Alignment of Figurative Language in Humans and Large Language Models}

\author{Samhita Bollepally \\
  \texttt{ samhita.bollepally@tamu.edu } \\\And
  Aurora Sloman-Moll \\
  \texttt{auroraslomanmoll@tamu.edu} \\
  \AND
  Takashi Yamauchi \\
  \texttt{takashi-yamauchi@tamu.edu} \\\And
  Neeka Motamedrasa \\
  \texttt{nmotamedrasa@gmail.com} \\}

\begin{document}
\maketitle

\begin{abstract}
Large language models (LLMs) often produce judgments similar to those of humans, but their interpretation of figurative language remains unclear. We compared human participants and four instruction-tuned LLMs (GPT-4, Gemma-2-9B, Llama-3.2, and Mistral-7B) on their ratings of 240 dialogue-based sentences spanning six linguistic traits: conventionality, sarcasm, funny, emotion, idiomacy, and Gen Z slang. While humans and LLMs were strongly aligned at the surface level, differences emerged in representational similarity patterns, particularly in idiomacy and Gen Z slang. GPT-4 showed the closest alignment to humans, though all models differed at the representational level, suggesting that similar outputs may arise from different underlying language representations. We also examined the relationship between model scale and representational similarity using the Gemma model series (Gemma-2-9B-IT, Gemma-3-27B-IT, and Gemma-4-31B-IT), and found that representational similarity patterns were largely unaffected by model size.

\end{abstract}

\section{Introduction}
Human language is the primary medium through which individuals convey thoughts, emotions, and subjective interpretations of the world. The meaning of language is shaped by contextual cues such as social environment, shared background knowledge, prior experiences, and tone. These contextual factors are important for interpreting pragmatically enriched forms of language, including humor, sarcasm, emotions, and slang. Yet, LLMs are known to generate fluent, human-like text despite lacking access to communicative signals such as contextual cues, facial expressions, vocal tones, or gestures. 

Comparing LLMs with humans across a broader range of tasks shows mixed strengths. \citep{Cai2024},  \citep{Karanikolas2023}, \citep{hu2024language} describe how LLMs balance natural language understanding and generating (NLU and NLG) like abilities in ways that echo aspects of human language, including phonology, grammatical constructions, and pragmatics, whereas studies by \citep{Alsajri2024}, \citep{Akter2023}, and \citep{Atox2024} show that ChatGPT excels at grammatical precision. At the same time, Gemini often performs better in contextual comprehension or reasoning. Together, these findings suggest that, although LLMs produce human-like responses \citep{Cuskley2024}, it remains unclear how well they capture human-like linguistic structures in domains that require pragmatic inference. \citep{Durt2023} note that human language is rooted in subjective experience and preconscious processes, which current LLMs lack. This perspective is reinforced by \citep{Dentella2024}, who show that instability in LLM responses on basic comprehension tasks indicates reliance on surface-level pattern matching rather than structured, contextually integrated semantic processing.

To address ongoing uncertainty about the depth of LLMs’ language interpretation, we examine whether their linguistic similarity representations resemble those of humans, using Representational Similarity Analysis (RSA) \cite{kriegeskorte2008representational, Kriegeskorte2013, yamauchi2025reading}. 

\subsection{Related Work}
A central question in current research is the extent to which large language models (LLMs) process language in ways comparable to humans. One line of work examines whether LLMs align with human judgments in pragmatic interpretation tasks. For example, \citep{Bojic2025} reports that GPT-4 can outperform humans on structured dialogue-based pragmatic benchmarks, but cautions that these gains rely on highly controlled datasets that may not reflect naturalistic language use. Similarly, \citep{Hu2023} shows that LLMs can make human-like choices in some zero-shot pragmatic tasks, yet systematically fail in contexts requiring theory of mind or sensitivity to speaker intent. These findings suggest that apparent alignment with human judgments may depend strongly on task structure.

Several studies also highlight persistent limitations in LLMs’ semantic and figurative reasoning, raising questions about whether surface-level agreement reflects deeper interpretive similarity. \citep{Shani2025} find that although LLMs can form broad conceptual categories resembling those of humans, they fail to capture fine-grained distinctions such as typicality. \citep{ye2025unveiling} report inconsistent metaphor comprehension, particularly under syntactic manipulations, suggesting fragility in how figurative meaning is internally represented. 

Recent work examines how language models process figurative expressions such as idioms, metaphors, sarcasm, and irony, and how computational methods can improve their handling of figurative language tasks. For example, \citep{jang2023figurative} find that although LLMs are increasingly competent at recognizing figurative patterns, they still rely on shallow lexical cues like keywords and local word co-occurrence patterns, rather than semantic relationships between words and syntactic cues as humans do. They suggest incorporating features such as syntactic structures associated with figurative constructions into training and evaluation benchmarks that explicitly capture human interpretive strategies for figurative language. Complementing this, \citep{chakrabarty2022flute} presents a dataset that improves figurative language understanding by adding human-written explanations of non-literal meanings. \citep{lai2024survey} highlights several technical improvements, such as incorporating semantic constraints to maintain metaphor coherence and applying reinforcement learning or controlled methods to generate figurative-style text. 

Additionally, studies on model interpretability motivate the examination of internal representations rather than behavioral accuracy. \citep{Poliak2025} argues that human comprehension relies on structural priors guiding meaning construction beyond surface word order, while \citep{Bayzayeva2025} and \citep{Preda2012} frame language as both cognition and communication, suggesting that LLMs optimized for prediction may not replicate human interpretive processes. According to \citep{Deacon1998}, symbols derive their significance from their systematic position within a network of other symbols, rather than just pointing to objects in the real world. Consequently, to compare human and LLM language comprehension, we must analyze the relational architecture of meaning \cite{Edelman1998, Deacon1998, Shepard1987, Yamauchi2005, Yamauchi2009}, in which concepts are defined by their mutual differences and systemic positions rather than by flat output metrics.

These perspectives motivate the use of Representational Similarity Analysis (RSA) to test whether humans and LLMs structure pragmatic and figurative language, such as sarcasm, idioms, and slang, in similar semantic spaces. Prior work shows partial alignment: \citep{Ogg2024} found that GPT-4 exhibits the closest representational similarity to humans. In contrast, \citep{Giallanza2024} demonstrated that human-like behavioral responses can still arise from divergent internal representational geometry, cautioning against equating surface-level agreement with shared understanding.

Together, these studies suggest that although LLMs may align with human judgments at the surface level, it remains unclear whether their underlying linguistic representations of figurative language mirror those of humans and how this alignment varies across models. The present work addresses this gap by jointly examining surface-level judgments and representational similarity structures across linguistic traits in multiple LLMs, enabling a systematic comparison of how closely different models process figurative and socially grounded language in human-like ways. It also investigates whether model scale is associated with differences in representational similarity patterns between humans and models.

\subsection{Overview}
\begin{figure}[h]
\centering
\includegraphics[width=0.6\columnwidth, height= 6cm, alt={The figure shows an example of how a sentence and questions were presented to humans to rate on a 10-point Likert scale}]{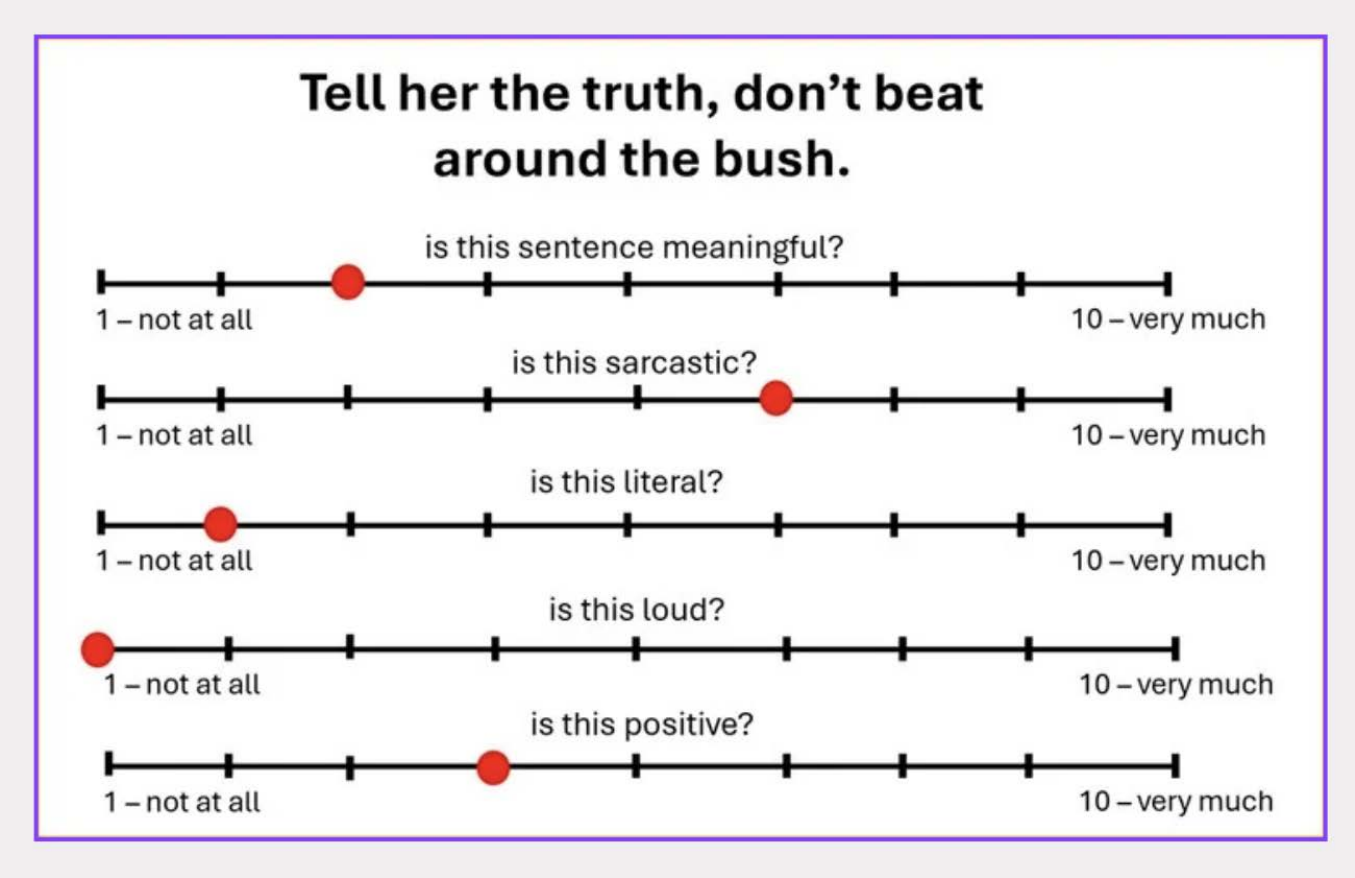}
\caption{The figure shows an example of how a sentence and questions were presented to humans to rate on a 10-point Likert scale.}
\label{fig: Human UI}
\end{figure}

This study examines how closely large language models (LLMs) align with human interpretations of figurative language by comparing both surface-level judgments and representational structures. Human participants and LLMs rated sentences spanning multiple semantic categories: conventional, idiomatic, emotional, funny, sarcastic, and Gen Z slang on a 10-point Likert scale using 40 expression-based questions (e.g., \textit{is this relevant to you?, Is this sarcastic?, Does this concern you?, etc}; Figure \ref{fig: Human UI}). The models evaluated were GPT-4, Gemma-2-9B-IT, Mistral-7B-Instruct-v0.3, and Llama-3.2-3B-Instruct. 

Surface-level similarity (SLS) was assessed by computing Pearson correlations between human and model ratings for each sentence category (Figure \ref{fig:SLS_workflow}). To examine representational similarity between linguistic traits, sentence-level representational spaces were constructed separately for humans and models by computing pairwise Euclidean distances between sentence rating vectors, yielding representational dissimilarity matrices that capture relative patterns of interpretation (Figure \ref{fig:RSA_workflow}). Representational Similarity Analysis (RSA) quantified alignment by correlating these matrices across humans and models, allowing comparison of interpretive structure independent of absolute rating values. Higher correlations indicate closer alignment in how interpretations are organized, whereas lower correlations reflect systematic differences in linguistic processing. 

Models in the Gemma series, including Gemma-2-9B-IT, Gemma-3-27B-IT, and Gemma-4-31B-IT, were tested to compare their judgments and representations with those of humans, aiming to investigate the association between model size and similarity representations.

\section{Method}
\subsection{Materials}
A total of 240 sentences were compiled, comprising 120 sensible dialogues commonly used in everyday communication and 120 non-sensible dialogues (see Table \ref{tab:appendix_all_sentences}). These sentences correspond to six categories: conventional, idiomatic, emotional, funny, sarcastic, and Gen Z slang, with each category containing 20 sensible and 20 non-sensible sentences. Sensible sentences contained one of these defining linguistic characteristics; in contrast, non-sensible sentences omitted the key characteristics that contribute to the linguistic trait (Table \ref{tab:sentences}). For example, \textit{“bite the bullet”} was classified as a sensible idiomatic sentence, whereas \textit{“eat the bullet,”} despite its structural similarity, was categorized as a non-sensible variant. Similarly, \textit{“the dog is outside”} was treated as a sensible conventional sentence, while \textit{“the outside is the dog”} was classified as its non-sensible counterpart. 

Each of the 240 sentences was paired with 40 interpretive questions designed to assess meaning, tone, and expressive quality of each sentence (\textit{e.g., loud, sentimental, concerning, positive, etc.}) (Table \ref{tab:questions}). This design yielded 9,600 sentence–question pairs. 

While the complete dataset was created by the authors, the majority of sentences and questions were curated from the internet, with a subset of reference sentences extracted from \citep{Rabagliati2018}. For a complete list, refer to Tables \ref{tab:appendix_all_sentences} and \ref{tab:questions}.

\subsection{Participants (Humans and Language Models)}

\textbf{\textit{Humans}}. A total of 235 undergraduate participants were recruited for course credit. Data from 14 participants were excluded due to incomplete responses, resulting in a final sample of 211 participants (137 female, 73 male, 1 non-binary; $M_{age}$ = 18.82), all of whom reported English as their primary spoken language.

\noindent\textbf{\textit{Language Models}}. Six instruction-tuned large language models were evaluated to examine language understanding and interpretation: GPT-4 (OpenAI), Llama-3.2-3B-Instruct (Meta), Gemma-2-9B-IT, Gemma-3-27B-IT, Gemma-4-31B-IT (Google), and Mistral-7B-Instruct-v0.3 (Mistral AI). The models developed by Meta, Google, and Mistral AI were accessed through the Hugging Face platform, while GPT-4 was accessed via an OpenAI API key. All models were pretrained on various publicly available datasets with varying parameter counts and then instruction-tuned to better align with user prompts (Table \ref{table:models_overview}). All models were evaluated using their default hyperparameters.

Language models were provided with the same dataset of 9,600 sentence–question pairs, identical to the human task, and evaluated under zero-shot prompting conditions. The prompts were formatted in a role–content structure, with input given in the user role. 

\begin{table*}[h]
\centering
\small
\setlength{\tabcolsep}{10pt}
\begin{tabular}{l l l}
\toprule
\textbf{Category} & \textbf{Type} & \textbf{Sentences} \\
\midrule
\multirow{2}{*}{Conventional} 
& Sensible & I adopted a dog today \\
& Non-sensible & The dog adopted me today \\
\midrule
\multirow{2}{*}{Idiomatic} 
& Sensible & I'm feeling under the weather \\
& Non-sensible & The weather is over me \\
\midrule
\multirow{2}{*}{Emotional} 
& Sensible & Max eagerly unwrapped a mysterious gift \\
& Non-sensible & Max quietly wrapped a mysterious gift \\
\midrule
\multirow{2}{*}{Funny} 
& Sensible & I used to be a baker because I kneaded dough \\
& Non-sensible & I used to be a baker, and had to knead dough \\
\midrule
\multirow{2}{*}{Sarcastic} 
& Sensible & I love your shirt, for now \\
& Non-sensible & I love your shirt \\
\midrule
\multirow{2}{*}{Gen Z Slang} 
& Sensible & This food is gas \\
& Non-sensible & The gas is food \\
\bottomrule
\end{tabular}
\caption{Sample sentences from each sentence category presented to humans and LLMs for judgment.}
\label{tab:sentences}
\end{table*}

\subsection{Procedure} \label{Procedure}
\textbf{\textit{Humans}}. Participants rated a subset of sentences drawn from the full set of 240 stimuli. Each sentence was paired with all 40 expressive questions and was rated on a 10-point Likert scale ranging from 1 (not likely) to 10 (very much). Participants were instructed to read each sentence carefully and provide ratings based on the displayed questions (Figure \ref{fig: Human UI}). Each participant evaluated approximately 32 sentences, randomly sampled from the full set and balanced across all sentence categories.

\noindent\textbf{\textit{Language Models}}. 
We employed two zero-shot prompting conditions. This design allowed us to assess the model's sensitivity to wording within the same prompt type and to examine whether explicit instructions led to systematic changes or improvements in rating patterns.

In \textbf{Study 1}, each model received a zero-shot prompt instructing it to read a sentence and rate it on a 10-point Likert scale in response to the accompanying question, the same as that given to the participants. Each of the 9,600 sentence–question pairs was presented individually, requiring the model to process the same prompt format repeatedly across all trials. The model outputs included both a numerical rating and a justification for each response, returned under the assistant role. This prompt was identical to the instructions given to human participants.

\textit{Prompt template 1= "Read the following sentence \{\}, and rate it on a Likert scale of 1-10 based on the given question \{\}."}

\textbf{Study 2} was designed to assess whether model–human alignment improved under refined prompt instructions. It followed the same procedure, with an additional constraint added to the zero-shot prompt. Models were instructed to (a) provide ratings strictly between 1 and 10, and (b) justify their choice in a single sentence. As in Study 1, all 9,600 inputs were processed.

\textit{Prompt template 2= "Read the following sentence \{\} and rate it on a Likert scale of 1-10 based on the given question \{\} and justify the rating in a sentence. Do not respond with a number other than [1, 10]."}

\textbf{Study 3} explores the relationship between model size and the representational similarity of linguistic traits when compared to humans. A series of Gemma-family models: Gemma-2-9B-IT, Gemma-3-27B-IT, and Gemma-4-31B-IT were prompted with template 2 to compare their judgments on surface-level and representational similarity with humans.

\subsection{Analysis}

\begin{figure*}[t]
\centering
\begin{subfigure}{0.48\textwidth}
    \centering
    \includegraphics[width=\linewidth, alt={Figure showing the frameowrk of the study}]{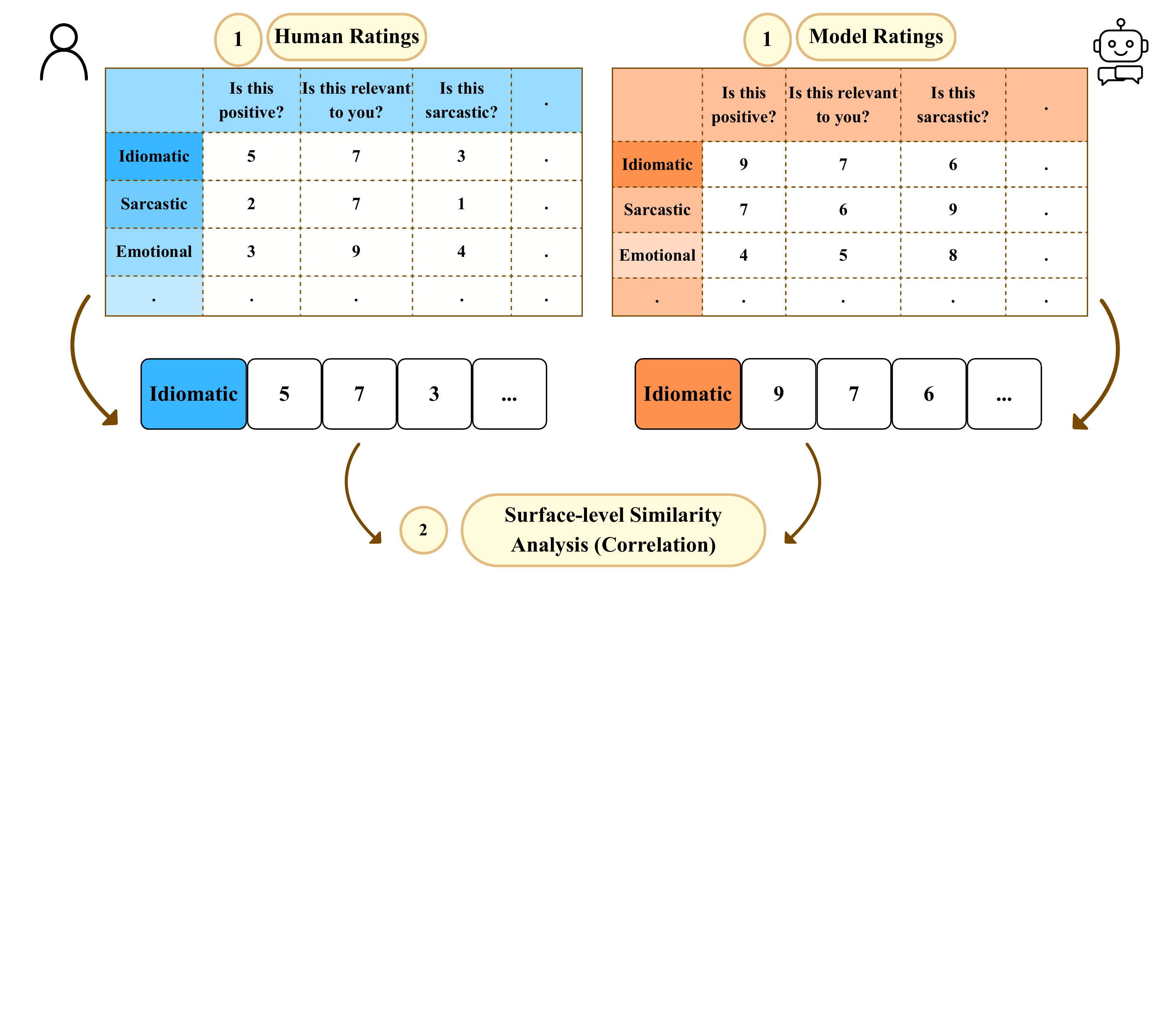}
    \caption{Surface-level similarity analysis}
    \label{fig:SLS_workflow}
\end{subfigure}
\vrule
\begin{subfigure}{0.48\textwidth}
    \centering
    \includegraphics[width=\linewidth]{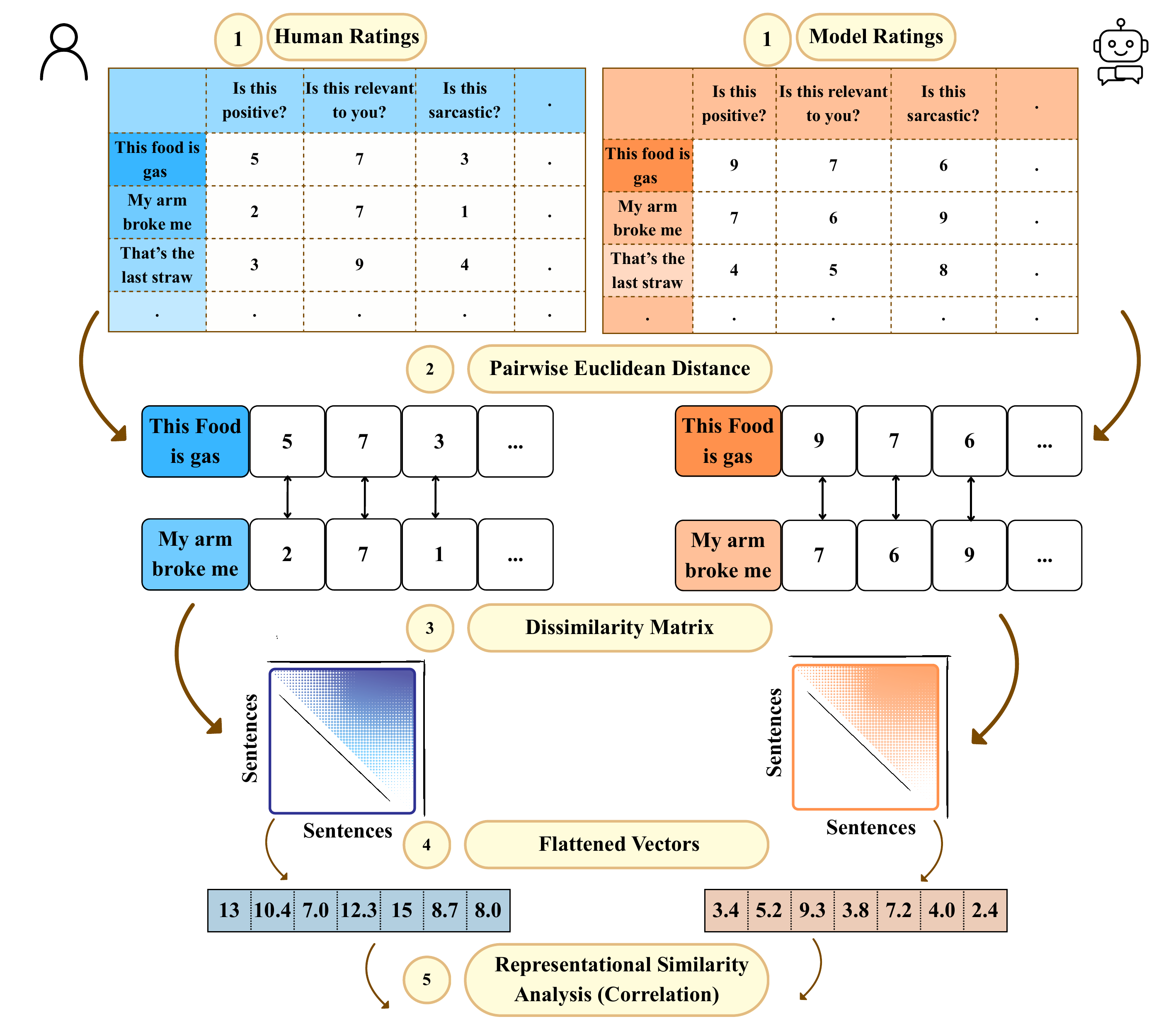}
    \caption{Representational similarity analysis}
    \label{fig:RSA_workflow}
\end{subfigure}
\caption{Human participants and LLMs rated all sentences on a 10-point Likert scale. In (a), ratings were aggregated by sentence category (e.g., idiomatic sentences) for humans and models, and Pearson correlations were computed between category-level mean ratings across the 40 dimensions (questions)($1\times40$) to assess surface-level alignment. In (b), sentence-level ratings across all categories (e.g., a Gen Z slang, \textit{“The food is gas”} and an idiomatic, \textit{“My arm broke me”}) sentences were used to compute pairwise Euclidean distances across 40-dimensional rating vectors, yielding a $240 \times240$ representational dissimilarity matrix. Representational alignment between humans and models was assessed by correlating the upper-triangular entries of their matrices.}
\label{fig:Workflow}
\end{figure*}

\textbf{Surface-level Similarity (SLS):} Surface-level similarity assesses the extent to which humans and models produce comparable rating patterns, independent of how those judgments are semantically organized. Pearson’s correlation coefficients were computed between human ratings and each LLM’s ratings across all questions within each category in both studies. This approach provides a direct measure of agreement at the output level between human and model language judgments (Figure \ref{fig:SLS_workflow}).

\noindent\textbf{Representational Similarity Analysis (RSA):} Humans and each model were examined in terms of how they organized sentences relative to one another. For each group, pairwise Euclidean distances were calculated between all sentence pairs using their ratings across all questions, resulting in a distance matrix that captured how similar or different the sentences were in the judgment space. Because the distance matrix was symmetrical, only the upper triangle containing the unique sentence pairs was retained. Correlation coefficients were then computed between the humans' and each model’s distance matrix, separately for each category of sentences (Figure \ref{fig:RSA_workflow}). This analysis quantified how similarly humans and models structured and represented relationships among sentences. Intuitively, sentences that were closer together in this space were interpreted more similarly, whereas sentences that were farther apart were interpreted more differently \cite{kriegeskorte2008representational, yamauchi2025reading}. For example, the analysis examines whether humans cluster conventional sentences separately from funny sentences, and whether LLMs exhibit similar clustering patterns, and to what extent.

\underline{\textit{Humans vs. Models.}} To evaluate similarities in language interpretation between humans and models, all sentences were clustered according to their defining linguistic characteristics (conventional, idiomatic, emotional, funny, sarcastic, and Gen Z slang), and both similarity analyses were conducted.

\underline{\textit{Humans vs. Humans.}} To establish a baseline for similarity analysis, human participants were randomly divided into two groups within each linguistic trait, and both SLS and RSA were computed between these groups. This comparison quantified the internal consistency of human judgments and provided an upper bound for evaluating model–human alignment.

\section{Results}
\begin{figure*}[h!]
\centering
\begin{subfigure}[t]{0.48\textwidth}
  \centering
  \includegraphics[width=\linewidth, alt={Bar plots comparing model–human and human–human correlations for sensible sentences in Study 1. Human–human correlations are near 1 for surface-level similarity but substantially lower for representational similarity. Model–human correlations are generally lower than the human baseline, particularly for representational similarity.}]{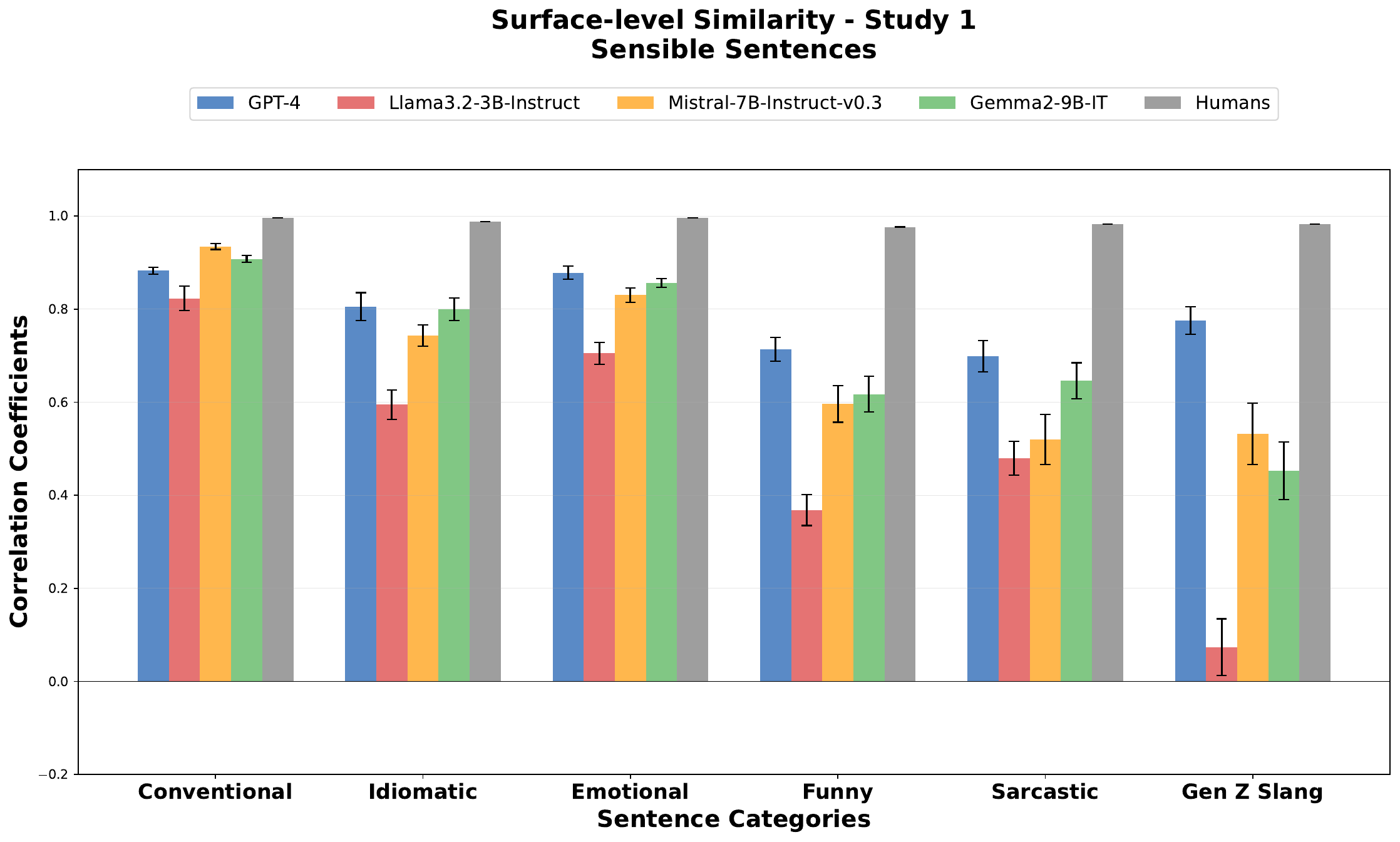}
  \caption{SLS — Sensible Sentences}
  \label{fig: SLS-S1-S}
\end{subfigure}
\hfill
\begin{subfigure}[t]{0.48\textwidth}
  \centering
  \includegraphics[width=\linewidth]{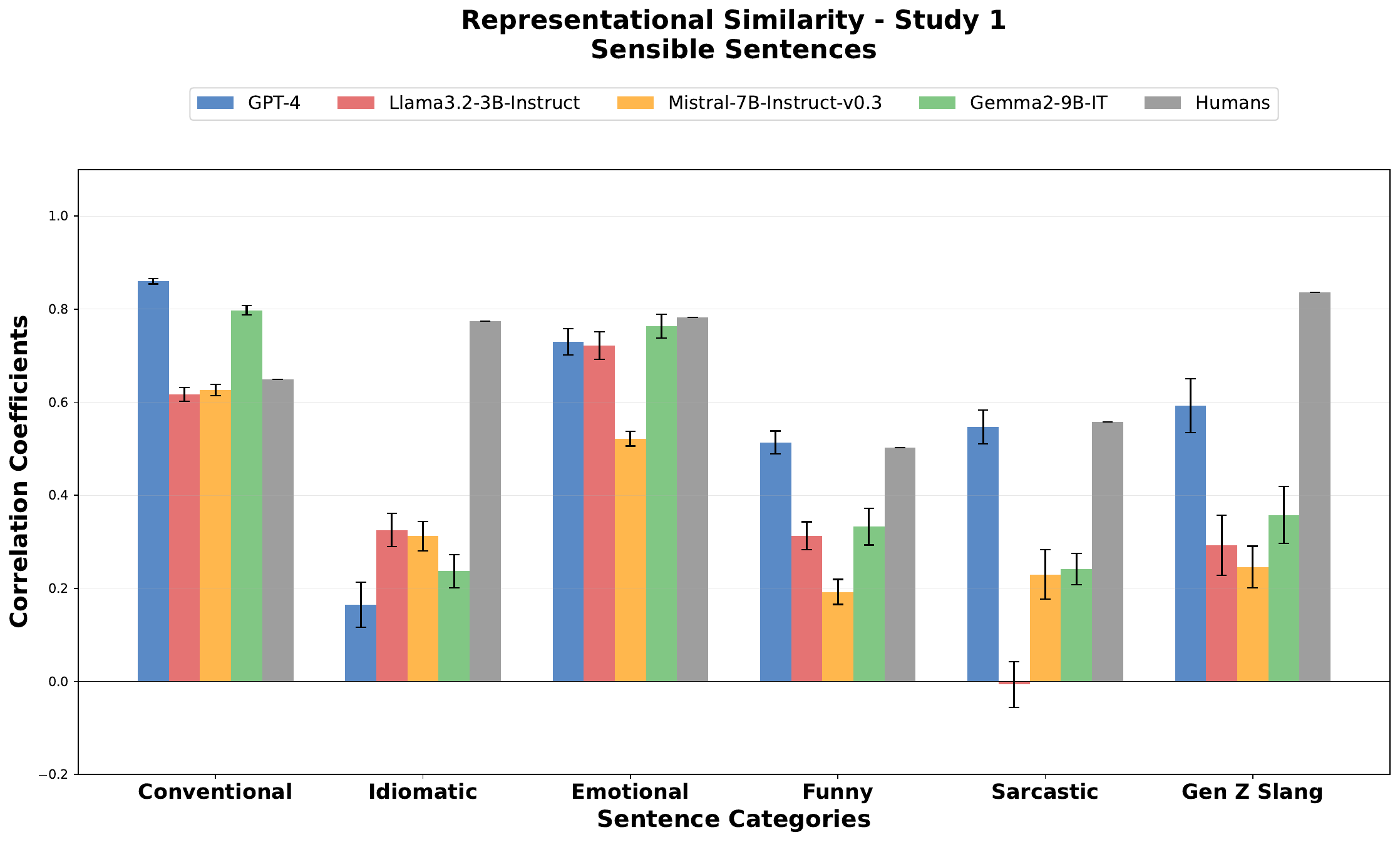}
  \caption{RSA — Sensible Sentences}
  \label{fig: RSA-S1-S}
\end{subfigure}
\caption{Surface-level similarity (SLS; a) and representational similarity analysis (RSA; b) between humans and language models for sensible sentences in \textbf{Study 1}. Colored bars represent model–human correlations, and gray bars represent the human–human baseline. Although models show relatively strong alignment with humans at the surface level (a), model–human correlations are substantially lower with RSA (b), indicating reduced alignment at the representational level relative to the human–human baseline.} 
\label{fig:DC}
\end{figure*}

\begin{figure*}[h!]
\centering
\begin{subfigure}[t]{0.48\textwidth}
  \centering
  \includegraphics[width=\linewidth, alt={Bar plots comparing model–human and human–human correlations for sensible sentences in Study 1. Human–human correlations are near 1 for surface-level similarity but substantially lower for representational similarity. Model–human correlations are generally lower than the human baseline, particularly for representational similarity.}]{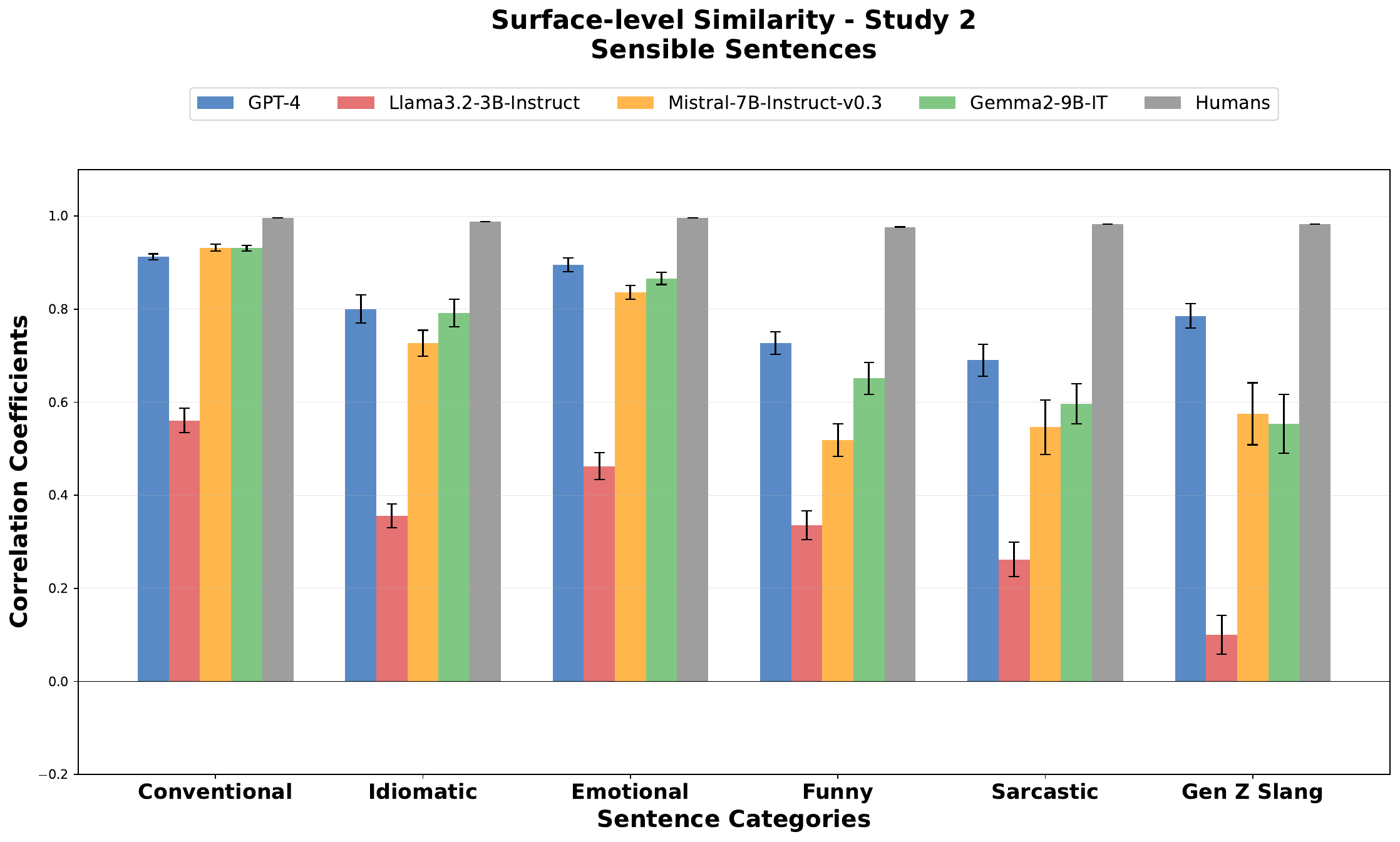}
  \caption{SLS — Sensible Sentences}
  \label{fig: SLS-S2-S}
\end{subfigure}
\hfill
\begin{subfigure}[t]{0.48\textwidth}
  \centering
  \includegraphics[width=\linewidth]{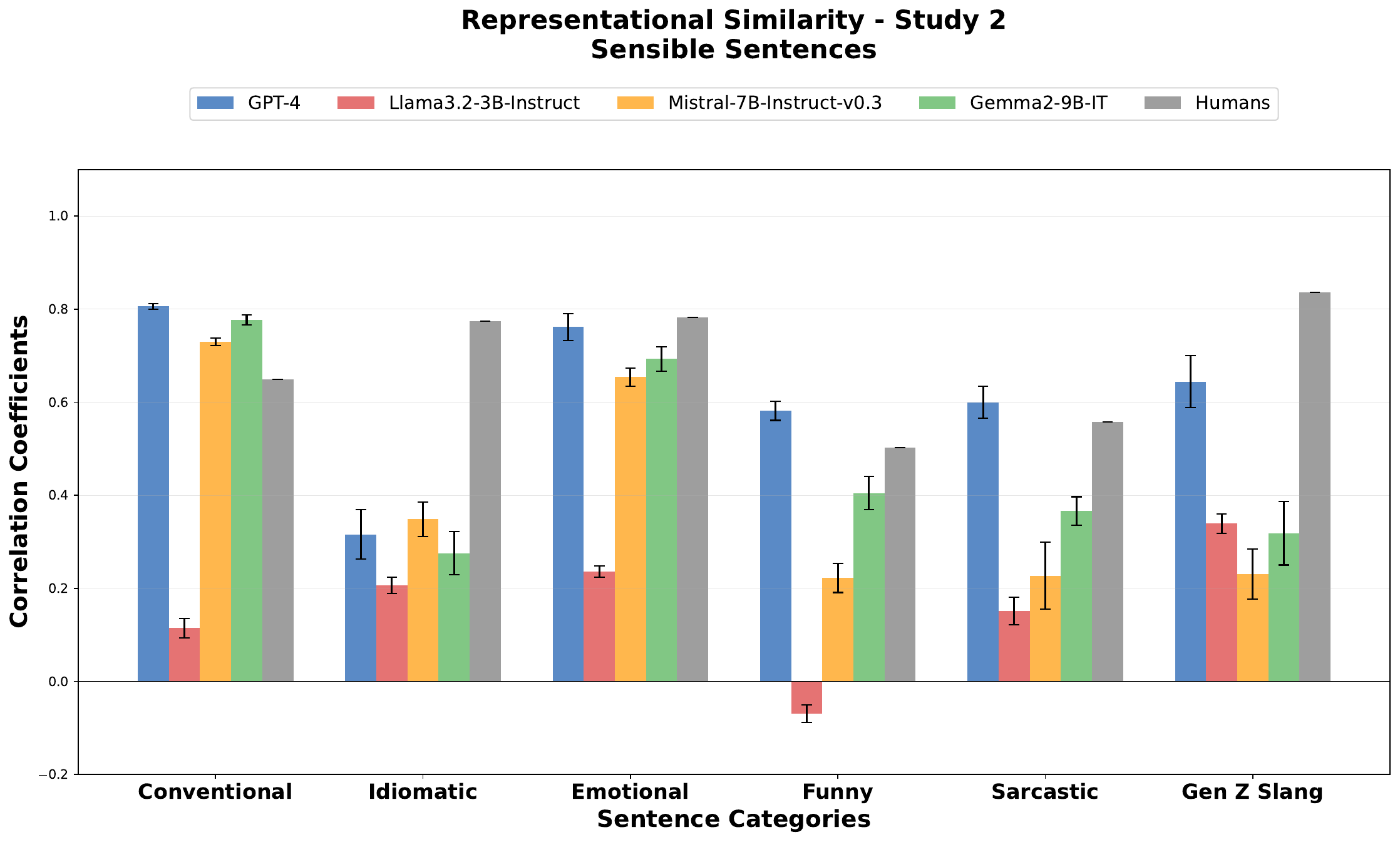}
  \caption{RSA — Sensible Sentences}
  \label{fig: RSA-S2-S}
\end{subfigure}
\caption{Figure 3: Surface-level similarity (SLS; a) and representational similarity analysis (RSA; b) between humans and language models for sensible sentences in \textbf{Study 2}. Colored bars represent model–human correlations, and gray bars represent the human–human baseline. Although models show relatively strong alignment with humans at the surface level (a), model–human correlations are substantially lower with RSA (b), indicating reduced alignment at the representational level relative to the human–human baseline.}
\label{fig:RSA}
\end{figure*}

Figures \ref{fig:DC} and \ref{fig:RSA} illustrate surface-level and Representational similarity results for sensible sentences, comparing human and LLM judgments and within human judgments across sentence categories in studies 1 and 2. The outputs of study 2 followed a more structured format, consisting of a numerical rating and a concise justification, compared to study 1.

\begin{figure*}[h]
\centering
\begin{subfigure}[t]{0.48\textwidth}
  \centering
  \includegraphics[width=\linewidth, alt={Bar plots comparing model–human and human–human correlations for sensible sentences in Study 1. Human–human correlations are near 1 for surface-level similarity but substantially lower for representational similarity. Model–human correlations are generally lower than the human baseline, particularly for representational similarity.}]{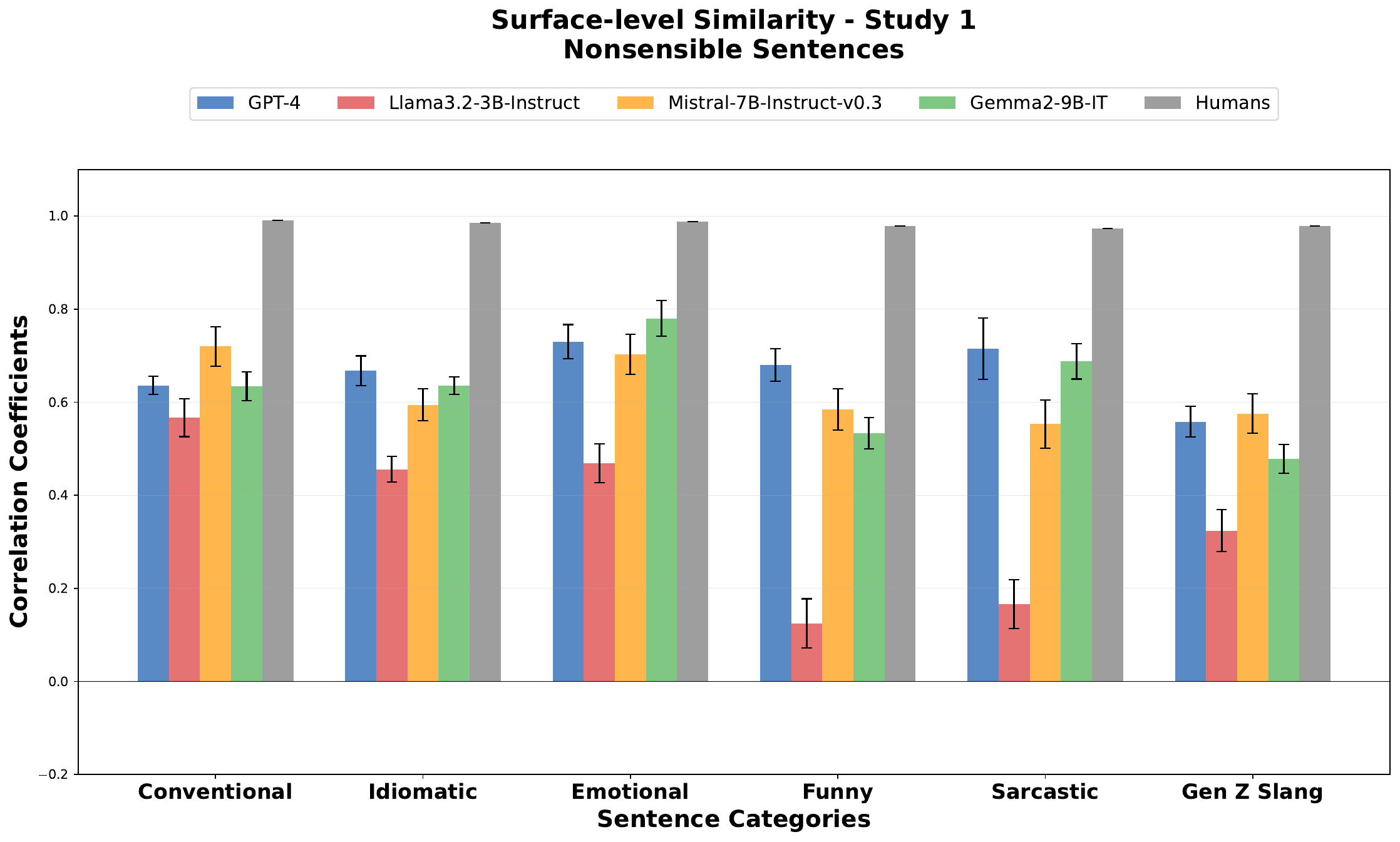}
  \caption{}
  \label{fig: SLS-S1-NS}
\end{subfigure}
\hfill
\begin{subfigure}[t]{0.48\textwidth}
  \centering
  \includegraphics[width=\linewidth]{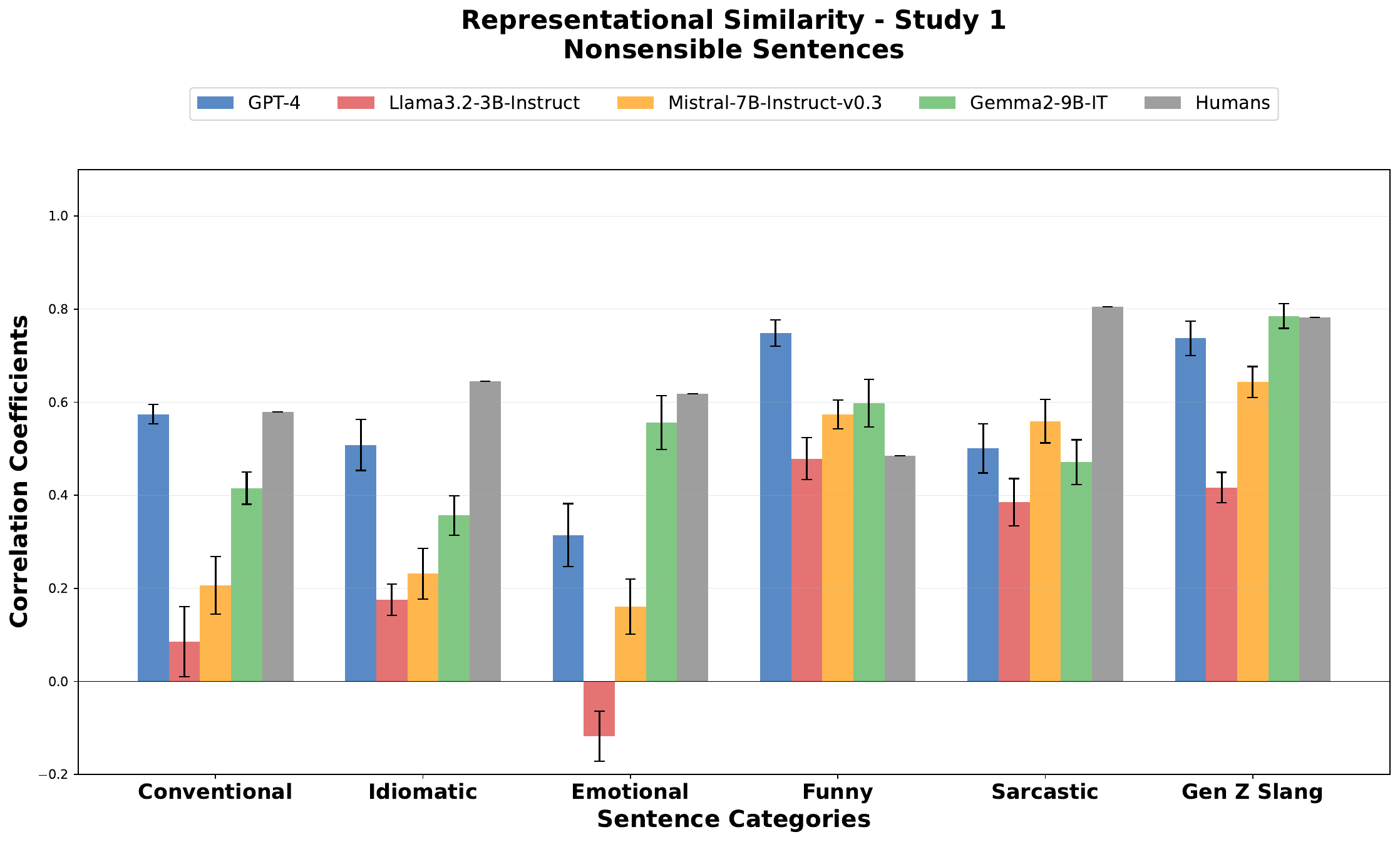}
  \caption{}
  \label{fig: RSA-S1-NS}
\end{subfigure}
\begin{subfigure}[t]{0.48\textwidth}
  \centering
  \includegraphics[width=\linewidth]{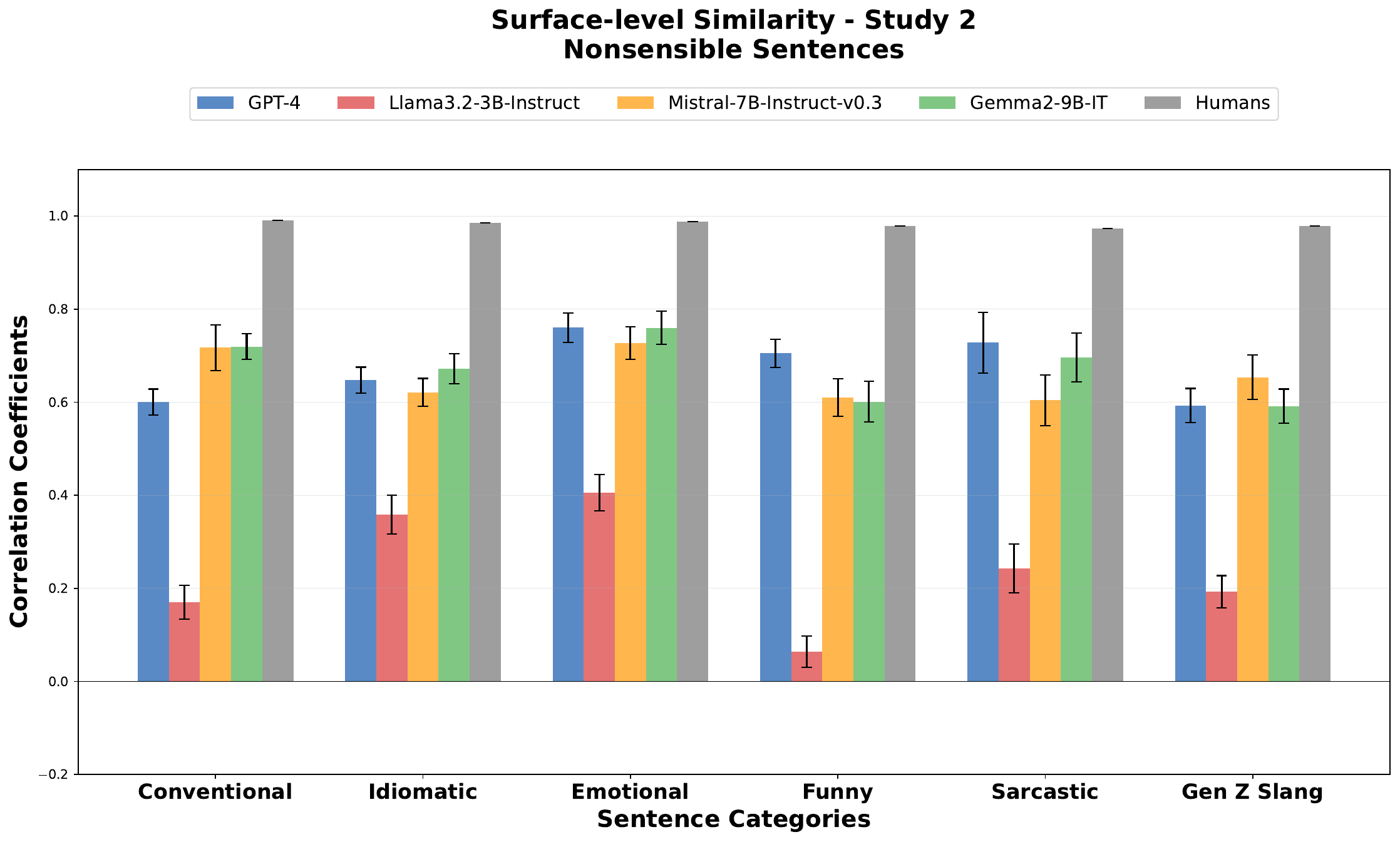}
  \caption{}
  \label{fig: SLS-S2-NS}
\end{subfigure}
\hfill
\begin{subfigure}[t]{0.48\textwidth}
  \centering
  \includegraphics[width=\linewidth]{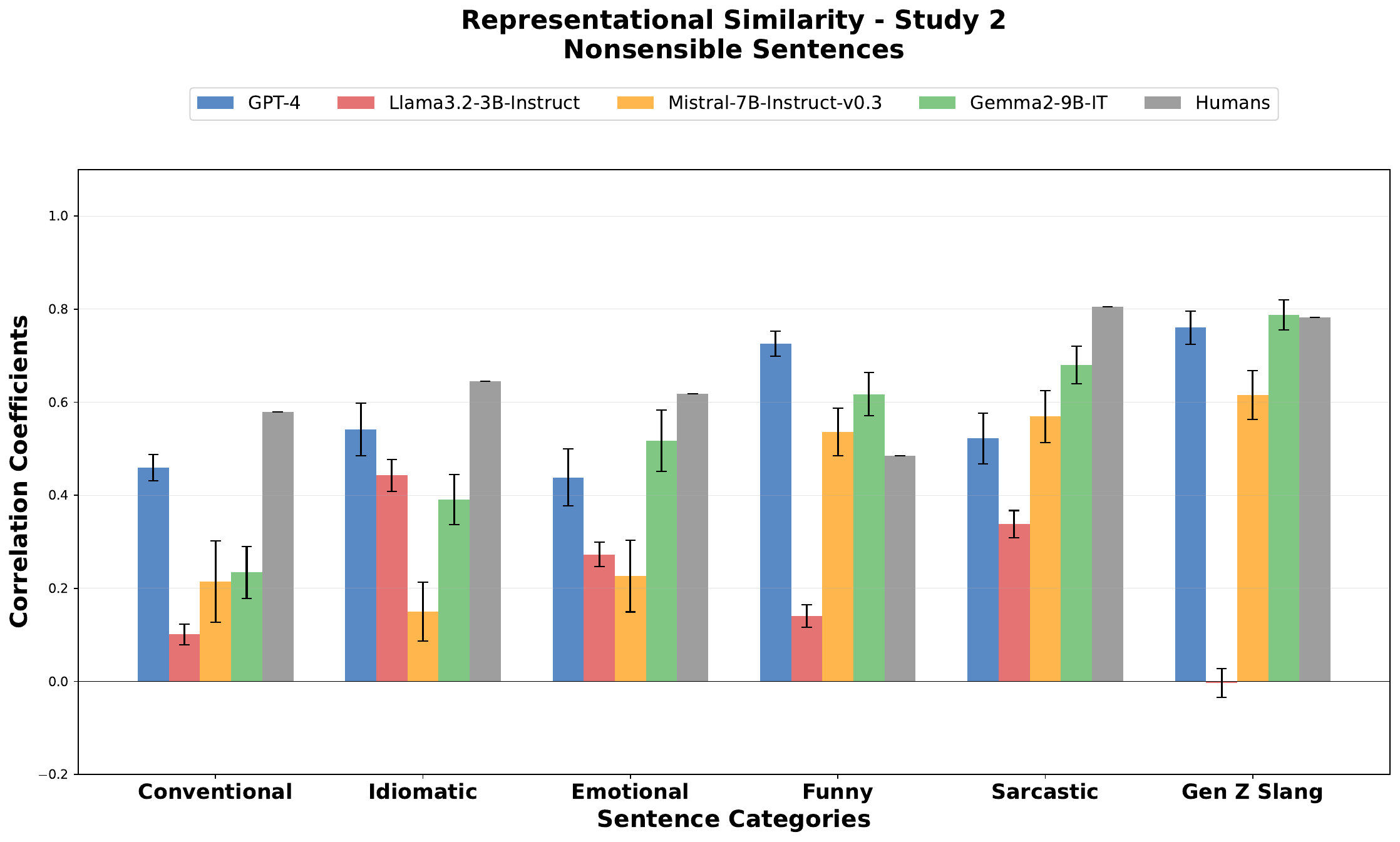}
  \caption{}
  \label{fig: RSA-S2-NS}
\end{subfigure}
\begin{subfigure}[t]{0.48\textwidth}
  \centering
  \includegraphics[width=\linewidth, alt={Bar plots comparing similarity across the Gemma model series, demonstrating that model size does not substantially influence the degree of similarity.}]{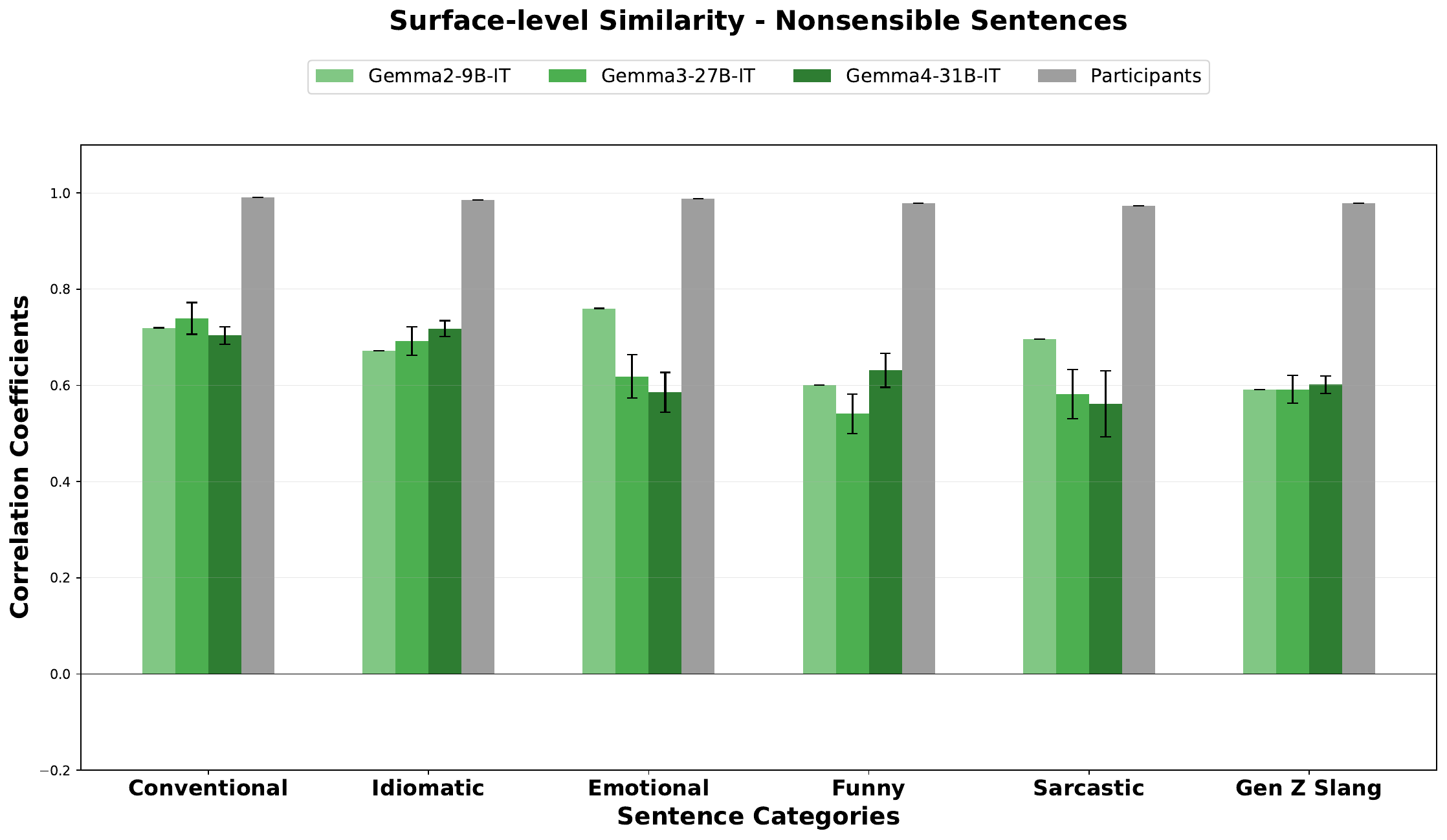}
  \caption{}
  \label{fig: G-DC-NS}
\end{subfigure}
\hfill
\begin{subfigure}[t]{0.48\textwidth}
  \centering
  \includegraphics[width=\linewidth]{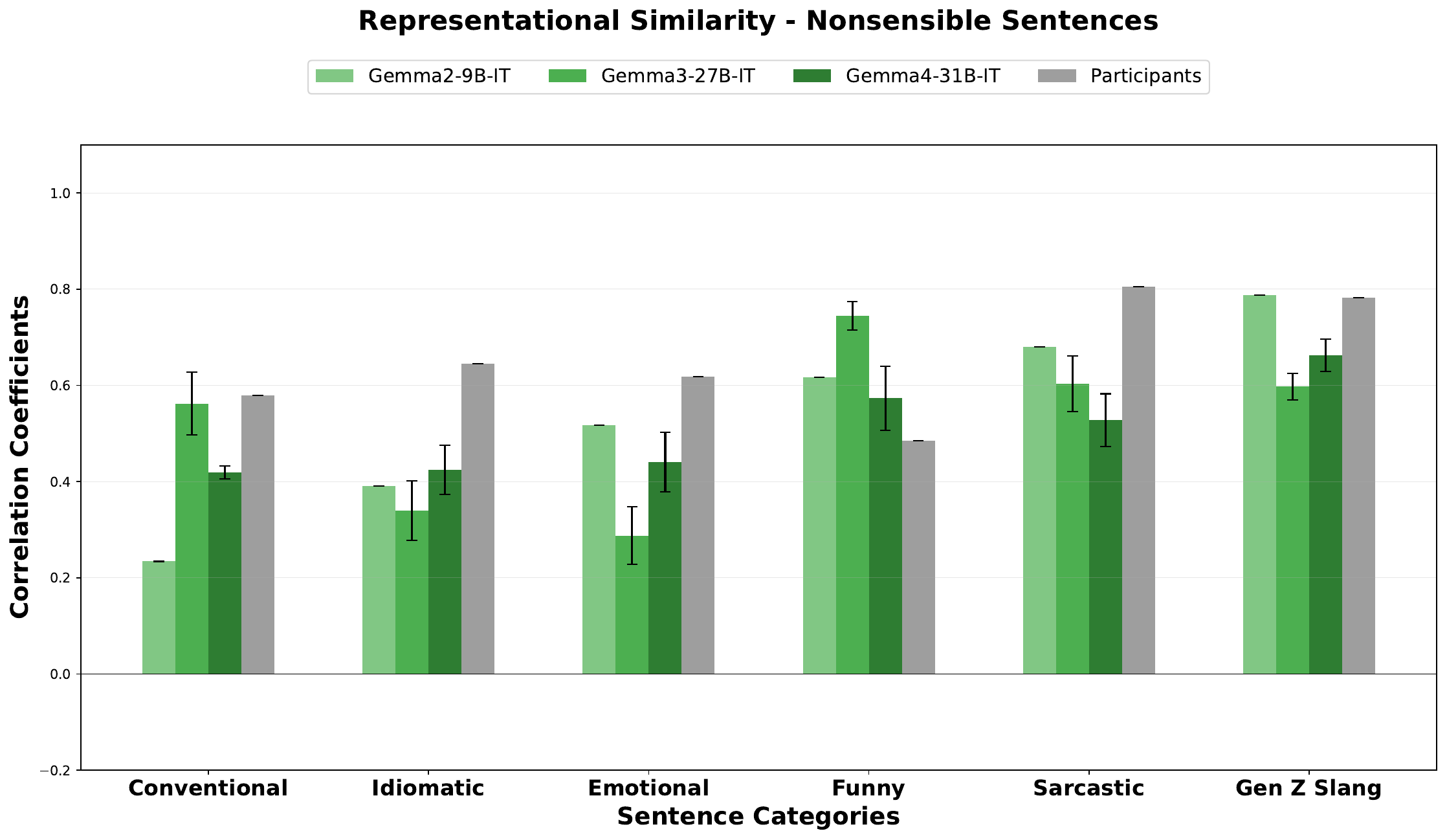}
  \caption{}
  \label{fig: G-RSA-NS}
\end{subfigure}
\caption{The Surface-level Similarity (a, c, e) and Representational Similarity (b, d, f) analyses among humans and models for non-sensible across study 1 and study 2. Figures (e) and (f) illustrate performance differences across the Gemma family models in study 2.}
\label{fig: NS}
\end{figure*}

\textbf{Surface Similarity Analysis}. Across both studies, GPT-4 shows the strongest and most consistent alignment with human judgments across sentence categories. GPT–human correlations typically range from \textit{r} = .60–.90 (Figures \ref{fig: SLS-S1-S}, \ref{fig: SLS-S2-S}). Gemma and Mistral exhibit moderate alignment with humans, with correlations generally ranging from \textit{r} = .50–.80, depending on sentence type. In contrast, Llama consistently shows the weakest alignment, particularly for figurative categories such as funny, sarcasm, and Gen Z slang, where correlations often fall below \textit{r} = .40  for study 2. All model-human surface-level similarity comparisons were highly significant (\textit{p} < .001) for both studies, while for non-sensible sentences, correlations remain lower but substantial (\textit{r} = .55–.75)(Figures \ref{fig: SLS-S1-NS}, \ref{fig: SLS-S2-NS})

Across all models, correlations were generally higher in study 2 than in study 1, indicating improved agreement between models and humans under revised instructions that encouraged a more structured response format consisting of a numerical rating and a concise justification. Across both studies, conventional and emotional sentences consistently yielded higher surface-level correlations than other linguistic traits.

\textbf{Representational Similarity Analysis}. For sensible sentences, RSA reveals moderate to high alignment between human and model representations across both studies. In study 1 (Figure \ref{fig: RSA-S1-S}), GPT exhibits the strongest representational similarity with humans, with correlations ranging approximately from \textit{r} = .50–.85 across categories, and the highest values observed for conventional and emotional sentences. Gemma and Mistral show moderate representational alignment (\textit{r} = .55–.80), again performing better on conventional and emotional sentences than on funny, sarcasm, or Gen Z slang. Llama demonstrates consistently low representational alignment with humans (\textit{r} = .10–.35), particularly for figurative categories like idiomatic (\textit{r} = .32, \textit{p} = .040), sarcastic (\textit{r} = .007, \textit{p} = .967), and Gen Z slang (\textit{r} = .29, \textit{p} = .215). However, non-sensible sentences remain challenging for all models (Figures \ref{fig: RSA-S1-NS}, \ref{fig: RSA-S2-NS}). Correlations between human groups remain high across all categories (\textit{r} = .95–1.00), indicating an upper bound for model performance.

Across both studies, representational similarity analyses reveal clear model-wise differences. In study 2, representational alignment for sensible sentences increases overall, with GPT maintaining high alignment with humans (\textit{r} = .60–.85), and Gemma and Mistral showing notable improvements (\textit{r} = .60–.80), though figurative categories consistently lag behind conventional and emotional sentences. Human–human RSA remains high and stable across studies (\textit{r} = .75–.85), suggesting a common representational space. For non-sensible sentences, model–human similarity is lower and more variable: GPT again shows the strongest alignment (\textit{r} = .45–.75), followed by Gemma and Mistral (\textit{r} = .40–.60), while Llama exhibits weak or near-zero correlations. Although study 2 shows improvements in sarcasm and Gen Z slang (\textit{r} = .55–.75), model–human alignment remains consistently lower than human–human similarity. Overall, a stable hierarchy emerges, with GPT most closely aligned with human representations across all sentence categories, followed by Gemma and Mistral, and Llama showing limited alignment.

\textbf{Human-model correlation comparisons.} All human–model surface-level correlation comparisons were highly significant (\textit{p} < .001), confirming robust differences between human judgments and model representations. After Bonferroni correction (\(\alpha\)= .0125), GPT showed the smallest mean gap from humans (\textit{r} = .25), followed by Gemma (\textit{r} = .30), Mistral (\textit{r} = .32), and Llama (\textit{r} = .62). Fisher’s Z tests revealed significant differences in 89 of 144 model comparisons, with the largest separations observed between GPT and Llama, particularly for Gen Z slang (study 1) and conventional and emotional content (study 2).

Descriptive distance analyses further showed that humans exhibited the tightest semantic clustering (Mean Euclidean Distance = 11.17), whereas all models operated at higher distances, indicating less compact semantic organization. Model performance varied by category: conventional sensible sentences showed the strongest correlations (up to \textit{r} = .93), while Gen Z slang and funny non-sensible sentences remained consistent failure cases, especially for Llama (\textit{r} = 0.06 – 0.10). Model behavior across studies was largely stable for GPT, Gemma, and Mistral (\textit{p} > .080), but Llama showed a marked decline from study 1 to study 2. Overall, these results demonstrate persistent differences between human and model semantic similarity representations, with GPT showing relatively closer alignment but all models struggling with figurative, humorous, and slang-based language.

\begin{figure*}[h]
\centering
\begin{subfigure}[t]{0.48\textwidth}
  \centering
  \includegraphics[width=\linewidth]{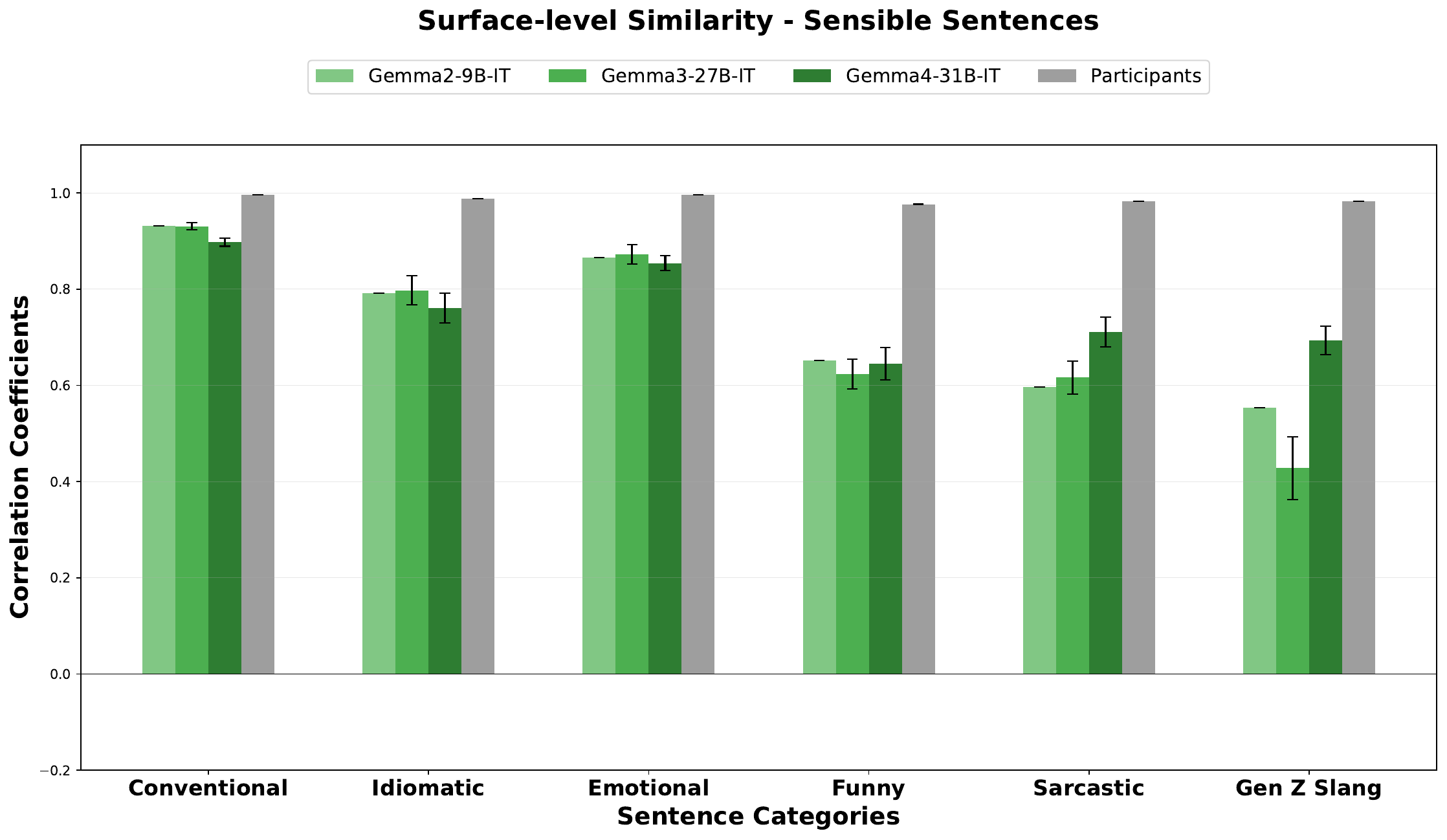}
  \caption{SLS — Sensible Sentences}
  \label{fig: G-SLS-S2-S}
\end{subfigure}
\hfill
\begin{subfigure}[t]{0.48\textwidth}
  \centering
  \includegraphics[width=\linewidth]{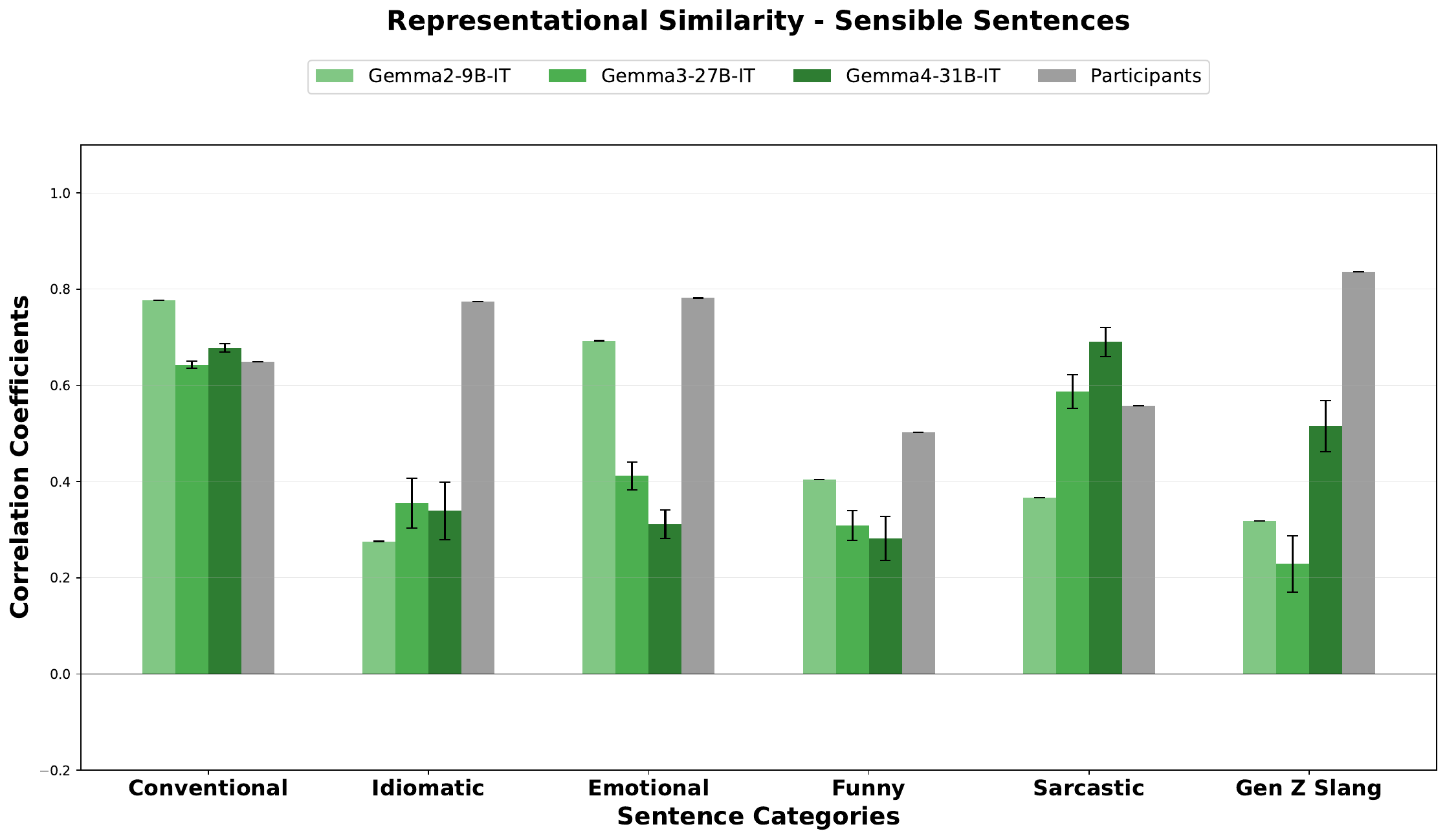}
  \caption{RSA — Sensible Sentences}
  \label{fig: G-RSA-S2-S}
\end{subfigure}
\vspace{1em}
\caption{The Surface-level similarity (a) and Representational Similarity (b) analyses among humans and Gemma models for sensible sentences across study 2.}
\label{fig:G}
\end{figure*}

\textbf{Effect of Model Size on Similarity Representations.}
In the Gemma series (Figure \ref{fig:G}), sentence-level similarity demonstrated a strong alignment with human judgment for conventional sentences (\textit{p} < .001), with correlation coefficients ranging from \textit{r} = .93 to .89. Both Gemma-3 and Gemma-4 exhibited improved performance on idiomatic sentences (\textit{r} = .76 to .79) and sarcastic sensible sentences, with Gemma-4 achieving the highest alignment at \textit{r} = .71.

In representational similarity analysis, Gemma-3 showed strong alignment for both conventional and sarcastic sensible sentences, with correlation coefficients of \textit{r} = 0.64 and \textit{r} = .59, respectively (\textit{p} < .001). However, it demonstrated weaker alignment for sensible content expressed in Gen Z slang, with a correlation of \textit{r} = .43, \textit{p} > .05. Gemma-4 showed significant improvements in pragmatic categories, particularly in sarcasm and Gen Z slang, with correlation coefficients of \textit{r} = .71 and \textit{r} = .69, respectively (\textit{p} < .001). In contrast, the idiomatic and funny-sensible categories remained relatively weak. Notably, there was a decline in representational similarity alignment for emotionally sensitive sentences across the models, with correlation coefficients ranging from \textit{r} = .70 to \textit{r} = .31 (\textit{p} < .05). 

Overall, all Gemma variants showed the best alignment with human understanding for conventional and emotional language. However, they struggled with figurative and non-sensible categories, particularly idiomatic expressions, humor, and slang, indicating ongoing challenges in modeling pragmatic and figurative language.

\section{Discussion}
At the surface level, LLMs, particularly GPT-4, approximated human judgments for nearly all categories of sentences. However, the persistent gap between model–human and human–human agreement for idiomatic and Gen Z slang suggests that surface-level alignment is fragile. These results support prior critiques that behavioral similarity can emerge from distributional pattern matching rather than stable semantic comprehension \citep{Cuskley2024, Durt2023}. 

Despite moderate to high correlations in surface-level similarity, the overall RSA revealed divergence between model and human representational similarity spaces. The contrast between emotional and idiomatic language further sharpens this result. Emotional sentences show relatively stronger representational alignment, likely because emotional meaning is often conveyed through explicit lexical and affective cues. In contrast, idiomatic expressions require non-compositional reinterpretation, where meaning cannot be derived directly from surface form. The weaker RSA alignment at the surface-level, observed for idioms, suggests that model representations privilege literal-form meaning over pragmatically inferred meaning, consistent with prior observations of limited pragmatic abstraction in LLMs \citep{Karanikolas2023, Dentella2024}.

Comparisons across models reveal a consistent hierarchy, with GPT-4 exhibiting stronger alignment with humans than other models at both surface and representational levels. GPT-4 fails to achieve human–human representational consistency, suggesting that improved alignment does not imply convergence toward a human-like semantic organization. The weaker representational alignment observed in the Llama and Mistral models is especially pronounced for pragmatically rich categories such as idiomatic, sarcastic, and Gen Z slang. This pattern suggests limitations in the models’ ability to encode context-sensitive cues within a single semantic representational space, and it varies across models.

One possible reason for the divergence in the alignment between the surface-level and representational similarities of models and humans could be model size. In an investigation of the Gemma series across different sizes, although there were inconsistencies in alignment quality between the Gemma models and humans in specific sentence categories, no consistent pattern of improved performance was observed in either analysis. Therefore, rather than attributing these differences solely to the model scale or size, presumably current architectures do not effectively enforce stable semantic clustering of pragmatic meaning. Since human interpretation of sarcasm and humor relies heavily on discourse context, speaker intent, and situational knowledge \citep{Hu2023, Shani2025}, the persistent representational gap suggests that humans and LLMs may rely on fundamentally different mechanisms for constructing meaning.

Consistent with prior work calling for improved evaluation frameworks for figurative language processing and understanding in language models \citep{banou2025systematic}, our approach collects responses to 40 interpretive questions for each item. This allows us to examine multiple dimensions of meaning construction and capture richer patterns of interpretation. By comparing these response structures between humans and LLMs, we can identify differences in representational similarities that would remain hidden in single-score evaluations. Future research could adopt hybrid evaluation approaches that integrate multiple complementary measures. In addition to behavioral metrics such as classification accuracy and rating-based judgments, evaluation can incorporate contextual reasoning assessments, where models explain or justify their interpretations of figurative expressions, as suggested by explanation-based approaches in the literature \citep{chakrabarty2022flute}. 

\subsection{Conclusion}
Using Representational Similarity Analysis (RSA), this study shows that although LLMs align closely with human judgments at the surface level, their underlying representational similarity structures only partially align with those of humans, particularly for figurative language traits such as idiomaticity, sarcasm, and Gen Z slang, where meaning extends beyond literal words. These results indicate that similar behavioral outputs can emerge from different underlying semantic representational organizations in LLMs that differ from those of humans. Additionally, the study finds that these representational similarity patterns are independent of model size.

\subsection{Limitations and Future Directions}
This study provides an initial step toward understanding representational similarities between humans and LLMs in interpreting figurative language, while also opening several promising directions for future research.

First, although the dataset of 240 sentences paired with 40 interpretive questions enabled detailed analysis, larger and more diverse datasets could strengthen the generalizability of the findings. Since the dataset was constructed based on the author's intuitions, future research may also benefit from standardized benchmark datasets and broader human validation to reduce potential bias.

Second, sentences were intentionally presented without additional contextual cues to isolate context-independent interpretation of linguistic traits across humans and LLMs. Although Likert-scale ratings simplify nuanced interpretations into numerical values, this controlled setup enabled direct comparison of surface-level and representational similarity patterns under identical conditions. Extending this framework with richer discourse context, conversational history, and open-ended reasoning measures may provide deeper insights into pragmatic language understanding.

Finally, the selected models represent only a subset of LLM families, limiting broader generalization across architectures. Further work can systematically examine how prompting strategies, training methods, architectural design, and model families influence representational similarity patterns and the structure of linguistic meaning.

\subsection{Study Materials}

The complete dataset is available in Appendices A and B (Tables \ref{tab:appendix_all_sentences} and \ref{tab:questions}). The complete behavioral data is available on our \href{https://github.com/SamhithaBollepally/Figrative_Language_Interpretation_in_LLMs.git}{\underline{GitHub Profile}}.

\subsection{Ethical Considerations}
This study was approved by the Institutional Review Board (IRB No. IRB2022-1126M). The ethical standards in the Declaration of Helsinki for research with human subjects were followed. Informed consent was obtained from all participants before the study. Data were collected from undergraduate participants recruited for course credit. Demographic information was collected in anonymized form and used only for aggregate analysis.
\clearpage

\bibliographystyle{unsrt}
\bibliography{references}
\clearpage

\onecolumn
\section*{Supplementary Materials}

\begin{table*}[h]
    \centering
    \small
    \setlength{\tabcolsep}{10pt}
    \begin{tabular}{l l l}
    \toprule
    \textbf{Model} & \textbf{Parameters} & \textbf{Primary Training Objective} \\
    \toprule
    GPT-4 & Not publicly disclosed &
    General-purpose language understanding and generation, \\
    & & including reasoning, dialogue, and instruction following \\
    \midrule
    Llama-3.2-3B-Instruct & 3 billion &
    Instruction-tuned language generation with an emphasis \\
    & & on efficiency and conversational alignment \\
    \midrule
    Gemma-2-9B-IT & 9 billion &
    Instruction-tuned model optimized for safe, \\
    & & high-quality text generation and reasoning \\
    \midrule
    Gemma-3-27B-IT & 27 billion & 
    Instruction-tuned model aimed for \\
    && conversational text understanding \\
    \midrule
    Gemma-4-31B-IT & 31 billion & 
    Instruction-tuned multimodal causal language modeling \\
    && for more advanced reasoning\\
    \midrule
    Mistral-7B-Instruct-v0.3 & 7 billion &
    Instruction-following language generation with \\
    & & strong performance on reasoning and dialogue tasks \\
    \bottomrule
    \end{tabular}
    \caption{Overview of Large Language Models Used in the studies}
    \label{table:models_overview}

\vspace{3em}
\begin{subtable}[t]{0.45\textwidth}
\centering
\begin{tabular}{lcc}
\toprule
Models & Study 1 & Study 2 \\
\toprule
GPT-4 & 0.74 & 0.76 \\
Llama-3.2-3B-Instruct & 0.31 & 0.30 \\
Mistral-7B-Instruct-v0.3 & 0.68 & 0.69 \\
Gemma-2-9B-IT & 0.68 & 0.71 \\
\bottomrule
\end{tabular}
\caption{Overall Surface-level similarity (SLS)}
\label{table:DC-overall}
\end{subtable}
\begin{subtable}[t]{0.45\textwidth}
\centering
\begin{tabular}{lcc}
\toprule
Models & Study 1 & Study 2 \\
\toprule
GPT-4 & 0.63 & 0.64 \\
Llama-3.2-3B-Instruct & 0.49 & 0.19 \\
Mistral-7B-Instruct-v0.3 & 0.49 & 0.49 \\
Gemma-2-9B-IT & 0.61 & 0.65 \\
\bottomrule
\end{tabular}
\caption{Overall Representational Similarity (RSA)}
\label{table:RSA-overall}
\end{subtable}
\caption{Overall surface-level similarity (left) and representational similarity (right) between humans and models across study 1 and study 2.}
\label{table:overall-comparison}
\end{table*}

\clearpage

\clearpage
\onecolumn
\appendix
\section{Complete Sentence Set}
\begin{longtable}{l l p{9cm}}
\toprule
\textbf{Category} & \textbf{Type} & \textbf{Sentence} \\
\midrule
\endfirsthead
\toprule
\textbf{Category} & \textbf{Type} & \textbf{Sentence} \\
\midrule
\endhead
\midrule
\multicolumn{3}{r}{\textit{Continued on next page}} \\
\endfoot
\endlastfoot
Conventional & Sensible & The dog is outside. \\
Conventional & Sensible & I opened the window for some fresh air. \\
Conventional & Sensible & My sister is playing the piano. \\
Conventional & Sensible & The fork is in the drawer. \\
Conventional & Sensible & I made coffee this morning. \\
Conventional & Sensible & She had a sandwich for lunch. \\
Conventional & Sensible & The cat drank the water. \\
Conventional & Sensible & He ironed his shirt. \\
Conventional & Sensible & Ally went to the mall. \\
Conventional & Sensible & I broke my arm. \\
Conventional & Sensible & He bought a computer at the store. \\
Conventional & Sensible & I microwaved a bowl of soup. \\
Conventional & Sensible & He folded the laundry this morning. \\
Conventional & Sensible & I adopted a dog today. \\
Conventional & Sensible & I washed the dishes after dinner. \\
Conventional & Sensible & The kids built a sandcastle on the beach. \\
Conventional & Sensible & There are 32 kids in the class. \\
Conventional & Sensible & The chef prepared a delicious meal. \\
Conventional & Sensible & The sun rises in the east and sets in the west. \\
Conventional & Sensible & Plants need sunlight for photosynthesis. \\
Idiomatic & Sensible & Do not cry over spilled milk. \\
Idiomatic & Sensible & Break a leg today! \\
Idiomatic & Sensible & I'll do that when pigs fly! \\
Idiomatic & Sensible & Just bite the bullet. \\
Idiomatic & Sensible & Tell her the truth, don't beat around the bush. \\
Idiomatic & Sensible & I was going to go to class, but it's raining cats and dogs. \\
Idiomatic & Sensible & She's going through a lot, cut her some slack. \\
Idiomatic & Sensible & That's the last straw. \\
Idiomatic & Sensible & I'm feeling under the weather. \\
Idiomatic & Sensible & Now we're back to square one. \\
Idiomatic & Sensible & The project is not rocket science. \\
Idiomatic & Sensible & It's time to hit the sack. \\
Idiomatic & Sensible & I think he's pulling my leg. \\
Idiomatic & Sensible & You're barking up the wrong tree. \\
Idiomatic & Sensible & This is a wild goose chase. \\
Idiomatic & Sensible & The homework is a piece of cake. \\
Idiomatic & Sensible & You can kill two birds with one stone. \\
Idiomatic & Sensible & Take it with a grain of salt. \\
Idiomatic & Sensible & The devil is in the details. \\
Idiomatic & Sensible & She's burning bridges. \\
Emotional & Sensible & Sam is so enthusiastic that he jumps out of the bed and begins to cheer. \\
Emotional & Sensible & Filled with pride, Emily beamed as she watched her child's first steps. \\
Emotional & Sensible & Bursting with excitement, Alex's heart raced when he opened his first college acceptance letter. \\
Emotional & Sensible & Susan cried for Lisa when she heard that her dog passed away. \\
Emotional & Sensible & She sobbed when she heard the news about her aunt. \\
Emotional & Sensible & Her jaw dropped when she saw who was at the door. \\
Emotional & Sensible & Max's face turned red and he clenched his fist. \\
Emotional & Sensible & Emily took a deep breath to calm her nerves before going on stage \\
Emotional & Sensible & Ava paced back and forth waiting for her exam results. \\
Emotional & Sensible & Max eagerly unwrapped a mysterious gift. \\
Emotional & Sensible & Sarah smiled warmly as she revisited cherished childhood photos \\
Emotional & Sensible & David sighed in frustration during a traffic jam. \\
Emotional & Sensible & Alex watched as the long-awaited event was canceled. \\
Emotional & Sensible & Alex's heart fluttered, anticipating his partner's arrival. \\
Emotional & Sensible & Mark teared up, moved by a heartfelt movie. \\
Emotional & Sensible & Jake twirled with joy at news of his dream job. \\
Emotional & Sensible & Mike angrily crumpled up his homework after getting a bad grade. \\
Emotional & Sensible & Sally smiled and shed a tear at her brother's graduation. \\
Emotional & Sensible & Sofia blushed and giggled when he told her a joke. \\
Emotional & Sensible & Her face turned white with shame when he read her diary. \\
Funny & Sensible & If money does not grow on trees, why do banks have branches? \\
Funny & Sensible & I used to be a baker because I kneaded dough. \\
Funny & Sensible & Parallel lines have so much in common, it's a shame they'll never meet. \\
Funny & Sensible & An apple a day keeps the doctor away, if you throw it hard enough. \\
Funny & Sensible & I sold my vacuum, all it was doing was gathering dust. \\
Funny & Sensible & Life is a bowl of soup, and I'm a fork. \\
Funny & Sensible & You're not adopted, but we've placed an ad. \\
Funny & Sensible & My jeans says eat a salad, but my heart says eat a pizza. \\
Funny & Sensible & My brain has too many tabs open. \\
Funny & Sensible & I have a food baby. \\
Funny & Sensible & Everyone has a right to be stupid, but some abuse that privilege. \\
Funny & Sensible & I'm on a seafood diet; I see food and I eat it. \\
Funny & Sensible & My doctor diagnosed me with too smart syndrome. \\
Funny & Sensible & I'm jealous of my parents because I'll never have a kid as cool as theirs. \\
Funny & Sensible & I get enough exercise pushing my luck. \\
Funny & Sensible & The first five days after the weekend are the hardest. \\
Funny & Sensible & The road to success is always under construction. \\
Funny & Sensible & Finally, spring is here! I am so thrilled I wet my plants. \\
Funny & Sensible & I'm never late, others are just too early. \\
Funny & Sensible & Hey dude, pick your brain, sometimes. \\
Sarcastic & Sensible & I love your shirt, for now. \\
Sarcastic & Sensible & Nice perfume. How long did you marinate in it? \\
Sarcastic & Sensible & You find your patience before I lose mine. \\
Sarcastic & Sensible & Light travels faster than sound. This is why some people appear bright until they speak. \\
Sarcastic & Sensible & If you find me offensive. Then I suggest you quit finding me. \\
Sarcastic & Sensible & Unless your name is Google, stop acting like you know everything. \\
Sarcastic & Sensible & I do not have the energy to pretend to like you today. \\
Sarcastic & Sensible & I love your shoes. Great shoes. What a surprise! \\
Sarcastic & Sensible & I envy your commitment to being fashionably late. \\
Sarcastic & Sensible & I'm so glad you posted another photo of your lunch. I was on the edge of my seat wondering what you ate today. \\
Sarcastic & Sensible & It's truly impressive how you make the easiest tasks look incredibly complicated. \\
Sarcastic & Sensible & Thank you for the unsolicited advice; it's exactly what I wanted to hear. \\
Sarcastic & Sensible & I love your new hair, I never thought I'd see chunky highlights in this decade. \\
Sarcastic & Sensible & I failed my exam, that's exactly what I needed this week! \\
Sarcastic & Sensible & I love working 40 hours a week for minimum wage! \\
Sarcastic & Sensible & I would agree with you, but then we would both be wrong. \\
Sarcastic & Sensible & I don't know why I have a headache; all I've done is forget to drink water and stare at a screen all day. \\
Sarcastic & Sensible & I am busy right now, can I ignore you some other time? \\
Sarcastic & Sensible & Life is good, you should get one. \\
Sarcastic & Sensible & If had a dollar for every smart thing you say, I would be poor. \\
Gen Z slang & Sensible & I got an A in beer pong. \\
Gen Z slang & Sensible & Have a seat. You can just move that laundry pile. \\
Gen Z slang & Sensible & She is so boujee with that Louis Vuitton bag. \\
Gen Z slang & Sensible & Those potato chips are bussin. \\
Gen Z slang & Sensible & He got so salty after I did not text back right away. \\
Gen Z slang & Sensible & Her speech was woke. \\
Gen Z slang & Sensible & I am so amped for tonight's basketball game! \\
Gen Z slang & Sensible & She will be OK after she blows off some steam. \\
Gen Z slang & Sensible & I bombed that exam. \\
Gen Z slang & Sensible & I heard the Greek life on campus is pretty fun. \\
Gen Z slang & Sensible & Wake up!. It's time to hit the books. \\
Gen Z slang & Sensible & Where is your girlfriend? I dunno. \\
Gen Z slang & Sensible & The movie was mid. \\
Gen Z slang & Sensible & That 50\% off sale at the campus can't be legit. \\
Gen Z slang & Sensible & I'm so bored, I need to hear some good tea. \\
Gen Z slang & Sensible & I'm beefing with my roommate. \\
Gen Z slang & Sensible & Why is your friend on this date with us? He is kind of a third wheel. \\
Gen Z slang & Sensible & This food is gas. \\
Gen Z slang & Sensible & I was gonna stay in tonight, but I don't wanna get FOMO. \\
Gen Z slang & Sensible & Hey Bro. You talk too much. Keep your shirt on. \\
Conventional & Non-sensible & The outside is the dog. \\
Conventional & Non-sensible & The fresh air opened the window. \\
Conventional & Non-sensible & The piano is playing my sister. \\
Conventional & Non-sensible & The drawer is in the fork. \\
Conventional & Non-sensible & I made morning this coffee. \\
Conventional & Non-sensible & He had a cat for lunch. \\
Conventional & Non-sensible & The cat drank the shirt. \\
Conventional & Non-sensible & His shirt ironed him. \\
Conventional & Non-sensible & The mall went to Ally. \\
Conventional & Non-sensible & My arm broke me. \\
Conventional & Non-sensible & He bought a store at the computer. \\
Conventional & Non-sensible & I microwaved a soup of bowl. \\
Conventional & Non-sensible & The laundry folded him this morning. \\
Conventional & Non-sensible & The dog adopted me today. \\
Conventional & Non-sensible & After dinner, the dishes washed me. \\
Conventional & Non-sensible & On the beach, the sandcastle built the kids. \\
Conventional & Non-sensible & There are 32 classes in the kids. \\
Conventional & Non-sensible & A delicious meal prepared the chef. \\
Conventional & Non-sensible & The east rises in the sun and the west sets in the moon. \\
Conventional & Non-sensible & Sunlight needs plants for photosynthesis. \\
Idiomatic & Non-sensible & Don't cry over an orange juice spill. \\
Idiomatic & Non-sensible & Break your arm today. \\
Idiomatic & Non-sensible & I'll do that when the pigs dance. \\
Idiomatic & Non-sensible & Eat the bullet. \\
Idiomatic & Non-sensible & Tell the truth and don't beat up the tree. \\
Idiomatic & Non-sensible & I was going to go to class, but it's raining fish and monkeys. \\
Idiomatic & Non-sensible & She's going through the slack, cut her a lot. \\
Idiomatic & Non-sensible & The straw is last. \\
Idiomatic & Non-sensible & The weather is over me. \\
Idiomatic & Non-sensible & Square one is back to me. \\
Idiomatic & Non-sensible & The rocket science is not the project. \\
Idiomatic & Non-sensible & It's time for the sack to hit me. \\
Idiomatic & Non-sensible & I think my leg is pulling him. \\
Idiomatic & Non-sensible & You're barking up the right flower. \\
Idiomatic & Non-sensible & This run for ducks is wild. \\
Idiomatic & Non-sensible & A piece of cake is the homework. \\
Idiomatic & Non-sensible & You can murder two chickens with one rock. \\
Idiomatic & Non-sensible & Take a grain of rice with it. \\
Idiomatic & Non-sensible & The details are in the devil. \\
Idiomatic & Non-sensible & The bridges burned her. \\
Emotional & Non-sensible & Sam is so sad that he jumps and cheers. \\
Emotional & Non-sensible & Filled with anger, Emily beamed as her child took her first steps. \\
Emotional & Non-sensible & Sadness burst Alex when he opened his college letters. \\
Emotional & Non-sensible & Susan weeped gleefully for Lisa when she heard that her dog passed away. \\
Emotional & Non-sensible & She heard the news about her aunt when she sobbed. \\
Emotional & Non-sensible & Her jaw clenched when she saw who was at the door. \\
Emotional & Non-sensible & Max's face turned white and he tighened his fist. \\
Emotional & Non-sensible & Emily took a deep breath to stir her nerves before going on stage \\
Emotional & Non-sensible & Ava ran around the exam results. \\
Emotional & Non-sensible & Max quietly wrapped a mysterious gift. \\
Emotional & Non-sensible & Sarah cried as she revisited cherished childhood photos \\
Emotional & Non-sensible & David smiled in frustration during a traffic jam. \\
Emotional & Non-sensible & Alex canceled the long-awaited event he watched. \\
Emotional & Non-sensible & Alex's heart sank, anticipating his partner's arrival. \\
Emotional & Non-sensible & The movie moved Mark and teared up. \\
Emotional & Non-sensible & Jake scratched his head at news of his dream job. \\
Emotional & Non-sensible & Mike crumpled up his homework after getting a nasty grade. \\
Emotional & Non-sensible & Sally scowled and shed a tea at her brother's graduation. \\
Emotional & Non-sensible & Sofia turned red when he told her a joke. \\
Emotional & Non-sensible & Her face turned down when he read her diary. \\
Funny & Non-sensible & If branches don't grow on banks, why do trees have money? \\
Funny & Non-sensible & I used to be a baker, and had to knead dough. \\
Funny & Non-sensible & Parallel lines will never meet. \\
Funny & Non-sensible & If you throw it at the doctor, the watermelon keeps the doctor away. \\
Funny & Non-sensible & I sold my vacuum that used to gather dust. \\
Funny & Non-sensible & I'm a bowl of soup with a fork. \\
Funny & Non-sensible & You're not adopted, but we've seen an ad. \\
Funny & Non-sensible & My jeans says buy a salad, my heart says sell a pizza \\
Funny & Non-sensible & My computer has too many tabs open. \\
Funny & Non-sensible & The baby is food. \\
Funny & Non-sensible & Everyone has a right to be stupid, but some are really stupid. \\
Funny & Non-sensible & I'm on a seafood diet; I eat fish every day \\
Funny & Non-sensible & My doctor diagnosed me with too nasty syndrome. \\
Funny & Non-sensible & I envy my parents because they have a cool car. \\
Funny & Non-sensible & I get enough exercise pushing my lucky chair. \\
Funny & Non-sensible & The first five days before the weekend are the hardest. \\
Funny & Non-sensible & The road to success is always under consideration \\
Funny & Non-sensible & Finally, spring is here! I am so thrilled I watered my plants. \\
Funny & Non-sensible & I'm never late, others are always late. \\
Funny & Non-sensible & Dude, sometimes your brain picks you. \\
Sarcastic & Non-sensible & I love your shirt. \\
Sarcastic & Non-sensible & Nice perfume. How long have you had that for? \\
Sarcastic & Non-sensible & The patience got lost and couldn't be found. \\
Sarcastic & Non-sensible & Light travels faster than sound. Some people appear bright until they smile. \\
Sarcastic & Non-sensible & If you find me offensive. Then I suggest you quit smoking. \\
Sarcastic & Non-sensible & Unless your name is Alphabet, stop acting like you know everything. \\
Sarcastic & Non-sensible & I do not like you today. \\
Sarcastic & Non-sensible & I love your shoes. Great shoes. \\
Sarcastic & Non-sensible & I envy your commitment to being fashionable. \\
Sarcastic & Non-sensible & I'm so glad you posted a photo of your lunch. I was wondering what you ate today. \\
Sarcastic & Non-sensible & It's truly impressive how you make the incredibly complicated tasks look easy. \\
Sarcastic & Non-sensible & Thank you for the advice; it's exactly what I needed. \\
Sarcastic & Non-sensible & I love your new hair, I never thought I'd see beautiful hair like that. \\
Sarcastic & Non-sensible & I needed to fail my exam this week. \\
Sarcastic & Non-sensible & I hate working 40 hours a week for minimum wage. \\
Sarcastic & Non-sensible & I would agree with you, we would both be right. \\
Sarcastic & Non-sensible & I don't know why I have a headache; all I've done is to drink a lot of water. \\
Sarcastic & Non-sensible & I'm ignoring right now, can I be busy some other time? \\
Sarcastic & Non-sensible & Life's one, you should get good. \\
Sarcastic & Non-sensible & If had a dollar for every smart thing you say, I would be rich. \\
Gen Z slang & Non-sensible & I got an F in Economics 101. \\
Gen Z slang & Non-sensible & Have a laundry pile, you can just move that seat. \\
Gen Z slang & Non-sensible & She is so boujee with that Wal Mart bag. \\
Gen Z slang & Non-sensible & Those bussin' are potato chips. \\
Gen Z slang & Non-sensible & He bought salt after I didn't text back. \\
Gen Z slang & Non-sensible & She woke her speech. \\
Gen Z slang & Non-sensible & Tonight's basketball game was exciting. \\
Gen Z slang & Non-sensible & She will be OK after she eats steamed rice. \\
Gen Z slang & Non-sensible & That exam bombed me. \\
Gen Z slang & Non-sensible & I heard the campus on greek life is pretty fun. \\
Gen Z slang & Non-sensible & Wake up! It's time to smack the books. \\
Gen Z slang & Non-sensible & Where is her boyfriend? No clue \\
Gen Z slang & Non-sensible & The mid was movie. \\
Gen Z slang & Non-sensible & That 50\% off sale at the campus is great. \\
Gen Z slang & Non-sensible & I need tea to be bored. \\
Gen Z slang & Non-sensible & My roommate is with beef. \\
Gen Z slang & Non-sensible & Why is your friend on this date with us? He's kind of a fourth tire. \\
Gen Z slang & Non-sensible & The gas is food. \\
Gen Z slang & Non-sensible & I don't wanna eat FOMO. \\
Gen Z slang & Non-sensible & Hey Bro. You smile too much. Keep your pants on. \\
\bottomrule
\caption{Complete list of all 240 sentences used in the study for the judgments from human participants and LLMs, organized by category and sentence type.}
\label{tab:appendix_all_sentences} 
\end{longtable}
\clearpage
\section{Interpretive Questions}
\begin{table}[h]
\centering
\small
\begin{tabular}{p{12cm}}
\toprule
\textbf{Questions} \\
\midrule
Is this meaningful? \\
Do you find this funny? \\
Is this surprising? \\
Is this relevant (to you)? \\
Is this grammatically correct? \\
Is this insulting? \\
Do you find this exciting? \\
Is this sentence dull? \\
Is this sarcastic? \\
Is this honest? \\
Does this sentence show attitude? \\
Is this ironic? \\
Is this objective? \\
Is this mocking you? \\
Is this frustrating? \\
Does this sound serious? \\
Is this literal? \\
Do you find this stupid? \\
Is this convincing? \\
Is this active? \\
Do you think this is dramatic? \\
Is this arrogant? \\
Is this sentimental? \\
Is this polite? \\
Is this demeaning to you? \\
Is this positive? \\
Is this informative? \\
Is this logical? \\
Is this intriguing? \\
Does this sound confident? \\
Is this ambiguous? \\
Is this concerning? \\
Is this provocative? \\
Does this sound like praise? \\
Does this sound loud? \\
Does this sound negative? \\
Is this sentence weird? \\
Do you find this suspicious? \\
Do you think this is melodramatic? \\
Does this sound sympathetic? \\
\bottomrule
\end{tabular}
\caption{List of 40 interpretive questions used to collect judgments from human participants and LLMs.}
\label{tab:questions}
\end{table}

\end{document}